\documentclass[10pt,twocolumn,letterpaper]{article}

\usepackage{cvpr} 

\usepackage{booktabs}
\usepackage{colortbl}
\usepackage{multirow}
\usepackage{colortbl}
\usepackage{xcolor}
\usepackage{graphicx}
\usepackage{makecell}
\usepackage{float}
\usepackage[accsupp]{axessibility}
\definecolor{darkgreen}{RGB}{0,150,0}
\definecolor{lightgreen}{RGB}{255,250,220}
\definecolor{cvprblue}{rgb}{0.21,0.49,0.74}
\usepackage[pagebackref,breaklinks,colorlinks,allcolors=cvprblue]{hyperref}

\def\paperID{10198} 
\def\confName{CVPR}
\def\confYear{2025}

\title{Generalized Diffusion Detector: Mining Robust Features from Diffusion Models for Domain-Generalized Detection}

\author{
Boyong He$^{1}$\footnotemark[1] ,
Yuxiang Ji$^{1}$\footnotemark[1] ,
Qianwen Ye$^{2}$,
Zhuoyue Tan$^{1}$, 
Liaoni Wu$^{1\,2}$\footnotemark[2] ,
\\ 
    $^1$\textit{Institute of Artifcial Intelligence, Xiamen University} \\
    $^2$\textit{School of Aerospace Engineering, Xiamen University} \\
    \texttt{\{boyonghe, yuxiangji, yeqianwen, tanzhuoyue\}@stu.xmu.edu.cn}\\
    \texttt{wuliaoni@xmu.edu.cn}\footnotemark[2]
  }

\begin{document}
\maketitle

\renewcommand{\thefootnote}{\fnsymbol{footnote}}
\footnotetext[1]{Equal contribution.} 
\footnotetext[2]{Corresponding author.}


%
\begin{abstract}
    Domain generalization (DG) for object detection aims to enhance detectors' performance in unseen scenarios. This task remains challenging due to complex variations in real-world applications. Recently, diffusion models have demonstrated remarkable capabilities in diverse scene generation, which inspires us to explore their potential for improving DG tasks. Instead of generating images, our method extracts multi-step intermediate features during the diffusion process to obtain domain-invariant features for generalized detection. Furthermore, we propose an efficient knowledge transfer framework that enables detectors to inherit the generalization capabilities of diffusion models through feature and object-level alignment, without increasing inference time. We conduct extensive experiments on six challenging DG benchmarks. The results demonstrate that our method achieves substantial improvements of 14.0\% mAP over existing DG approaches across different domains and corruption types. Notably, our method even outperforms most domain adaptation methods without accessing any target domain data. Moreover, the diffusion-guided detectors show consistent improvements of 15.9\% mAP on average compared to the baseline. Our work aims to present an effective approach for domain-generalized detection and provide potential insights for robust visual recognition in real-world scenarios. The code is available at \href{https://github.com/heboyong/Generalized-Diffusion-Detector}{Generalized Diffusion Detector}
\end{abstract}

\begin{figure*}[!h]
    \centering
    \includegraphics[width=0.95\textwidth]{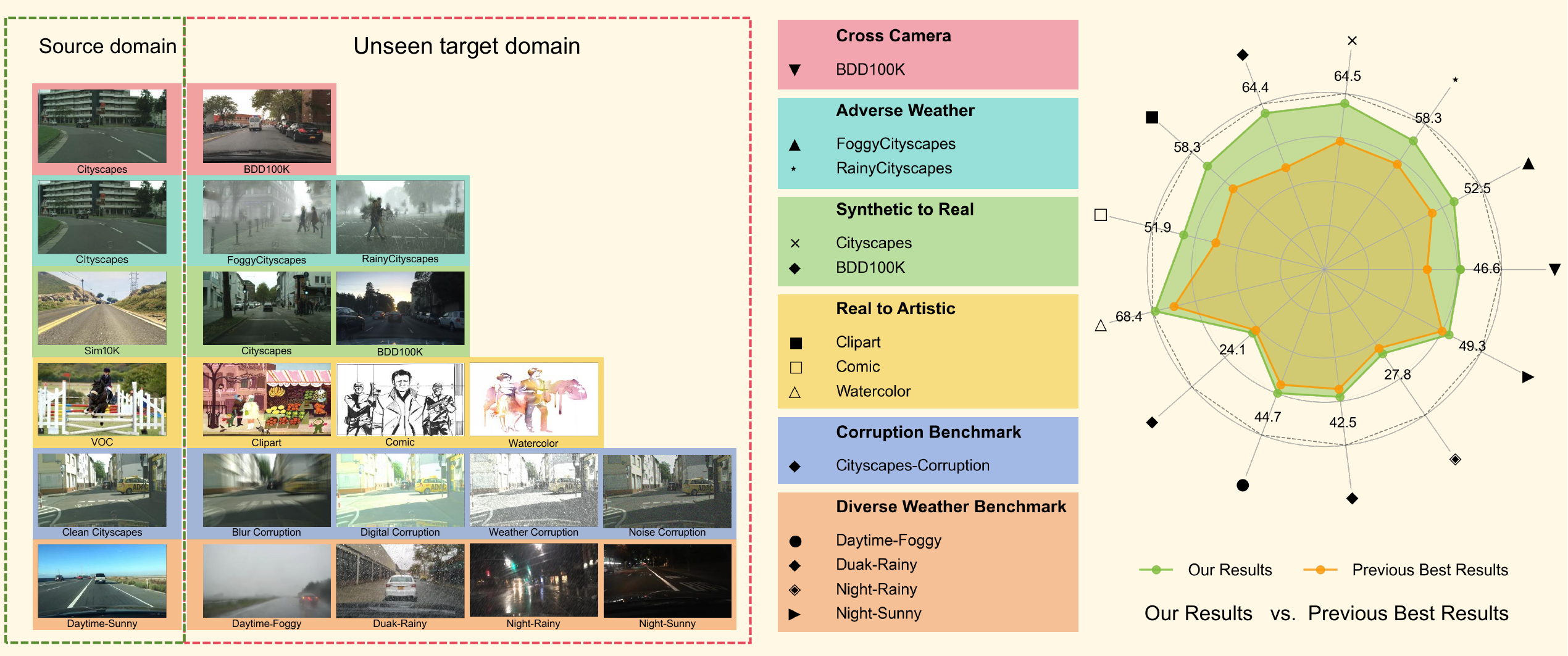}
    \caption{\textbf{Left:} Six DG benchmarks used in our paper. The sample images demonstrate substantial domain shifts between source and target distributions. \textbf{Right:} Compared with previous SOTA methods, our approach achieves superior performance across all 13 datasets from six benchmarks. }
    \vspace{-10pt}
    \label{fig:chart}
\end{figure*}

\section{Introduction}
\label{sec:intro}

Object detection stands as a fundamental task in computer vision and has achieved remarkable technological breakthroughs in recent years. Most object detection methods, including CNN~\cite{faster-rcnn,yolov3,lin2017retinanet,tian2020fcos} and transformer-based~\cite{carion2020Detr,zhu2021deformable} detectors assume consistent distributions between training and testing data. However, in practice, detectors face significant challenges from domain shifts and environmental variations, and the performance often deteriorates substantially when deployed in unseen scenarios~\cite{chen2018da-faster,dg}.

Domain generalization~\cite{episodic,adversarial,supervisions} and adaptation~\cite{hsu2020epm,AT,deng2023HT,cao2023cmt} methods have been developed to address these challenges. Mainstream approaches include semi-supervised learning with pseudo-labels~\cite{htcn,AT,deng2023HT,cao2023cmt}, feature alignment via adversarial training~\cite{chen2018da-faster,saito2019swda,crda,AT}, and style transfer techniques~\cite{gan,cyclegan}. However, these adaptation methods require target domain data during training, limiting their practical applications. This has led to increased interest in domain generalization through data augmentation~\cite{mixstyle,stylemix,Diversification,oamix}, adversarial training~\cite{dg_adver,localized}, and meta-learning~\cite{metareg,vibdg}. While recent advances in foundation models like ClipGap~\cite{clip_gap} have shown promising results, building robust detectors remains challenging.

Inspired by the remarkable capabilities of diffusion models~\cite{ho2020ddpm,song2020ddim,sd1} in handling visual variations, we propose to leverage them for domain-generalized detection. We extract and fuse multi-timestep intermediate features during the diffusion process to construct a diffusion-based detector that learns domain-invariant representations. However, directly applying these models introduces significant computational overhead due to the multi-step feature extraction process, limiting their practical deployment.

To address this limitation while preserving their strong generalization advantages, we develop an efficient knowledge transfer framework that enables lightweight detectors to inherit capabilities from diffusion-based detectors. Specifically, our framework consists of feature-level alignment using correlation-based matching and object-level alignment through shared region proposals, allowing conventional detectors to learn both domain-invariant features and robust detection capabilities. Through feature and object-level alignment, conventional detectors can achieve improved generalization without increasing inference time. Our work pioneers the application of diffusion models in domain-generalized detection, demonstrating their potential in enhancing detector generalization through knowledge distillation.

We conduct comprehensive experiments on six challenging DG benchmarks as shown in Fig.~\ref{fig:chart}: Cross Camera, Adverse Weather, Synthetic to Real, Real to Artistic, Diverse Weather Benchmark, Corruption Benchmark. Experimental results demonstrate that our diffusion-based detector achieves consistent improvements across these benchmarks, with average performance gains of \{\textbf{18.6, 15.0, 27.2, 16.4, 2.3, 4.7}\}\% mAP compared to previous methods, even outperforming most domain adaptation methods that have access to target domain data. Moreover, through our proposed feature-level and object-level learning framework, diffusion-guided detectors obtain significant improvements of \{\textbf{20.8, 21.4, 24.6, 9.9, 5.3, 13.8}\}\% mAP compared to their baselines. These results validate the effectiveness of leveraging diffusion models for domain-generalized detection and provide a promising direction for building robust detectors in real-world scenarios.

The main contributions of this work can be summarized as follows:
\begin{itemize}
    \item This work introduces diffusion models into domain-generalized detection. The inherent denoising mechanism and powerful representation capabilities of diffusion models are utilized to extract domain-invariant features for robust detection.
    
    \item To address the computational overhead, we propose a simple yet effective knowledge transfer framework. This framework enables detectors to inherit strong generalization capabilities through feature-level alignment and object-level learning, maintaining efficient inference time.
    
    \item Comprehensive experiments on six DG benchmarks demonstrate significant improvements over previous approaches in various scenarios. The findings provide valuable insights for robust visual recognition tasks.
\end{itemize}

\section{Related Work}
\label{sec:related work}
\subsection{Domain generalization for object detection}
Domain adaptation methods for object detection focus on adversarial feature alignment~\cite{chen2018da-faster,saito2019swda,crda,AT} and consistency-based learning with pseudo labels~\cite{htcn,AT,deng2023HT,cao2023cmt}. However, these approaches inherently require target domain data during training. Domain generalization methods have been explored through data augmentation~\cite{mixstyle,stylemix,randaugment,Diversification}, adversarial training~\cite{dg_adver,localized}, and meta-learning~\cite{metareg,vibdg} to enhance model robustness through style transfer and domain shift simulation. Recent works~\cite{zhou2022mga,gnas,oamix,Diversification,ufr} extend these strategies to domain-generalized detection through multi-view learning, specialized augmentation and causal learning. Additionally, ClipGap~\cite{clip_gap} demonstrates the potential of foundation models by leveraging CLIP~\cite{clip}. These advances inspire us to explore diffusion models as foundation models, harnessing their inherent generalization capabilities for domain-generalized detection.

\subsection{Diffusion models and applications}
Recent studies demonstrate that diffusion models~\cite{ho2020ddpm,song2020ddim,sd1,dalle,imagen,sd3} not only excel in image generation but also exhibit unique advantages in representation learning~\cite{ddpmseg,xu2023odise,ddt}. The noise-adding and denoising mechanism enables effective handling of visual perturbations like noise, blur, and illumination changes~\cite{hyperfeatures,tang2023emergent}. These properties suggest the potential of diffusion models in addressing domain generalization challenges. Specifically, the intermediate features during diffusion contain rich semantic information~\cite{hyperfeatures}, while the denoising process naturally builds robustness against various perturbations~\cite{tang2023emergent}. Recent works like~\cite{ddpmseg,ddt,ours_diff} further demonstrate the effectiveness of diffusion-based representations in various vision tasks. These promising properties motivate us to leverage diffusion models for domain-generalized detection, and further inspire us to explore transferring their superior capabilities to other detectors.

\section{Method}
\label{sec:method}

\subsection{Overview}
\begin{figure*}[t]
    \centering
    \includegraphics[width=1.0\textwidth]{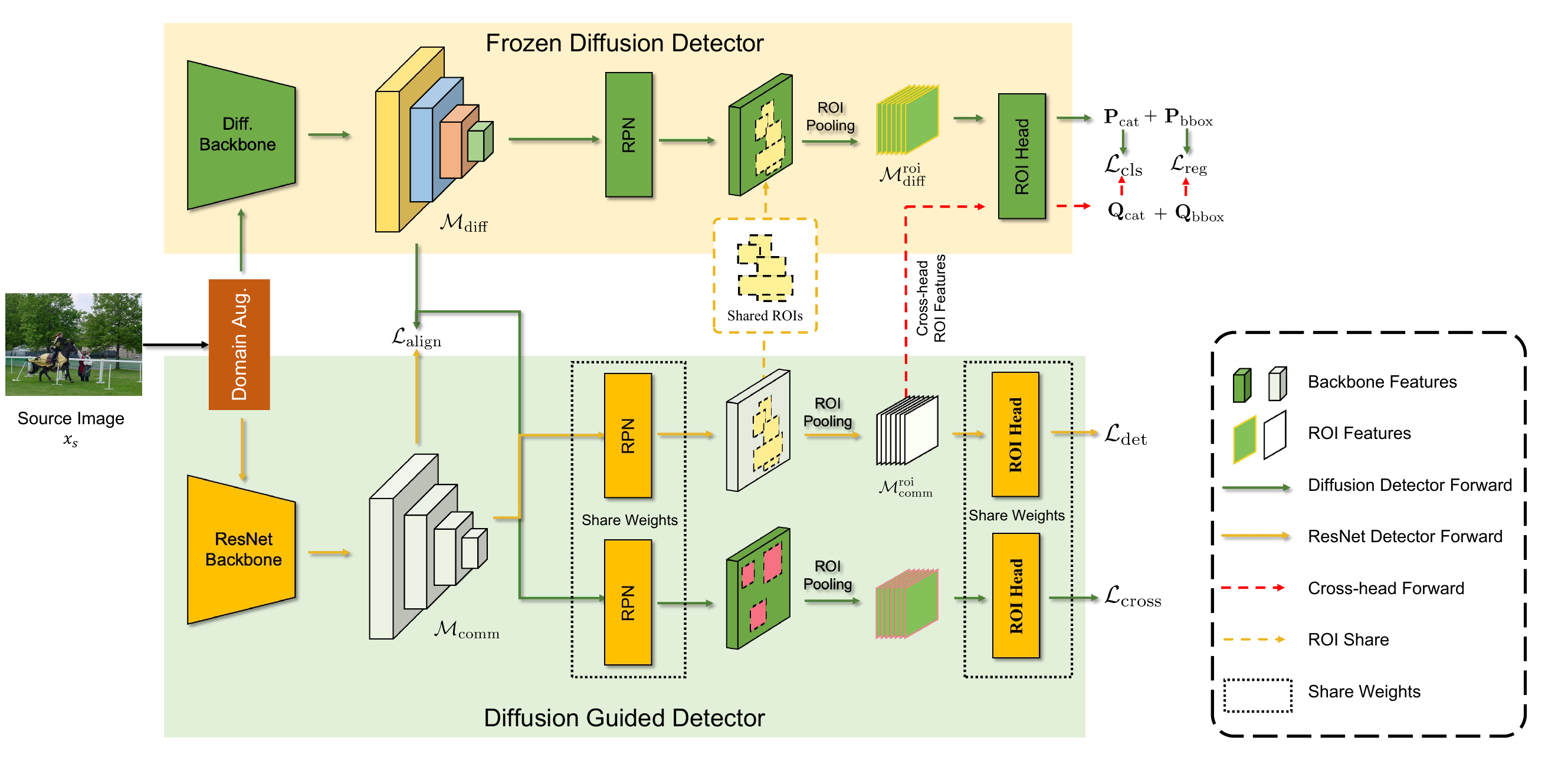}
    \caption{Overview of the proposed methods. The framework consists of a frozen diffusion detector (\textbf{top}) and a trainable ResNet-based detector (\textbf{bottom}). Knowledge transfer is achieved through feature-level alignment ($\mathcal{L}_{\text{align}}$, $\mathcal{L}_{\text{cross}}$) and object-level prediction alignment using shared RoIs ($\mathcal{L}_{\text{cls}}$ and $\mathcal{L}_{\text{reg}}$).}
    \label{fig:framework}
    \vspace{-8pt}
\end{figure*}

In this section, we introduce our approach for domain generalization detection. The proposed method leverages a diffusion model as feature extractor to learn domain-robust representations through its iterative process. A two-level alignment mechanism is designed to transfer knowledge from the diffusion model to standard detectors: feature-level alignment for domain-invariant patterns and object-level alignment for accurate detection across domains. The overall framework is illustrated in Fig.~\ref{fig:framework}.

In the following subsections, we first present the problem formulation and introduce the diffusion model basics in Sec.~\ref{sec:Preliminaries}. The feature extraction and fusion strategy from diffusion models is described in Sec.~\ref{sec:diffusion detector}. Our two-level alignment approach, including feature-level alignment (Sec.~\ref{sec:feature level}) and object-level alignment (Sec.~\ref{sec:object level}), guides standard detectors to learn robust representations while maintaining accurate detection. The overall training objectives integrating detection losses with alignment constraints are detailed in Sec.~\ref{sec:objective}.

\subsection{Preliminaries}
\label{sec:Preliminaries}
\noindent\textbf{DG for detection:} Let $\mathcal{S}=\left\{\mathbf{x}_s^i, \mathbf{y}_s^i\right\}_{i=1}^{N_s}$ denote the source domain with $N_s$ labeled samples, where $\mathbf{x}_s^i$ represents an image and $\mathbf{y}_s^i$ represents the corresponding bounding box annotations. Let $\mathcal{T}=\{\mathbf{x}_t^j\}_{j=1}^{N_t}$ denote the target domain with $N_t$ unlabeled images. The source and target samples are drawn from different distributions $P_\mathcal{S}$ and $P_\mathcal{T}$, with discrepancies in image distributions (e.g., style, scene), label distributions (e.g., shapes, density), and sample sizes. The goal is to learn robust representations from the labeled source domain that generalize well to the unseen target domain.

\noindent\textbf{Diffusion Model:} The diffusion process gradually converts source domain images $\mathbf{x}_s^i$ into pure noise through a fixed Markov chain of $T$ steps. At each time step $t \in [0,T]$, Gaussian noise is progressively added to obtain the noisy samples $\mathbf{x}_t$. The forward process is formulated as:
\begin{equation}
  \mathbf{x}_t=\sqrt{\bar{\alpha}_t} \mathbf{x}_0+\sqrt{1-\bar{\alpha}_t} \boldsymbol{\epsilon}
\end{equation}
where $\boldsymbol{\epsilon} \sim \mathcal{N}(0, \mathbf{I})$, and $\bar{\alpha}_t$ represents the cumulative product of noise schedule $\{\alpha_t\}_{t=1}^T$ controlling the noise magnitude. During training, a neural network $\mathcal{F}_{\theta}$ learns to estimate the added noise from noisy observations $\mathbf{x}_t$ conditioned on time step $t$. The network is optimized to minimize the discrepancy between predicted and actual noise. At inference, the model reverses this process by iteratively denoising pure noise through refinement steps until obtaining the final output $\hat{\mathbf{x}}_0$.

\subsection{Multi-timestep Feature Extraction and Fusion}
\label{sec:feature extract}
\label{sec:diffusion detector}

\begin{figure}[!h]
    \centering
    \includegraphics[width=0.5\textwidth]{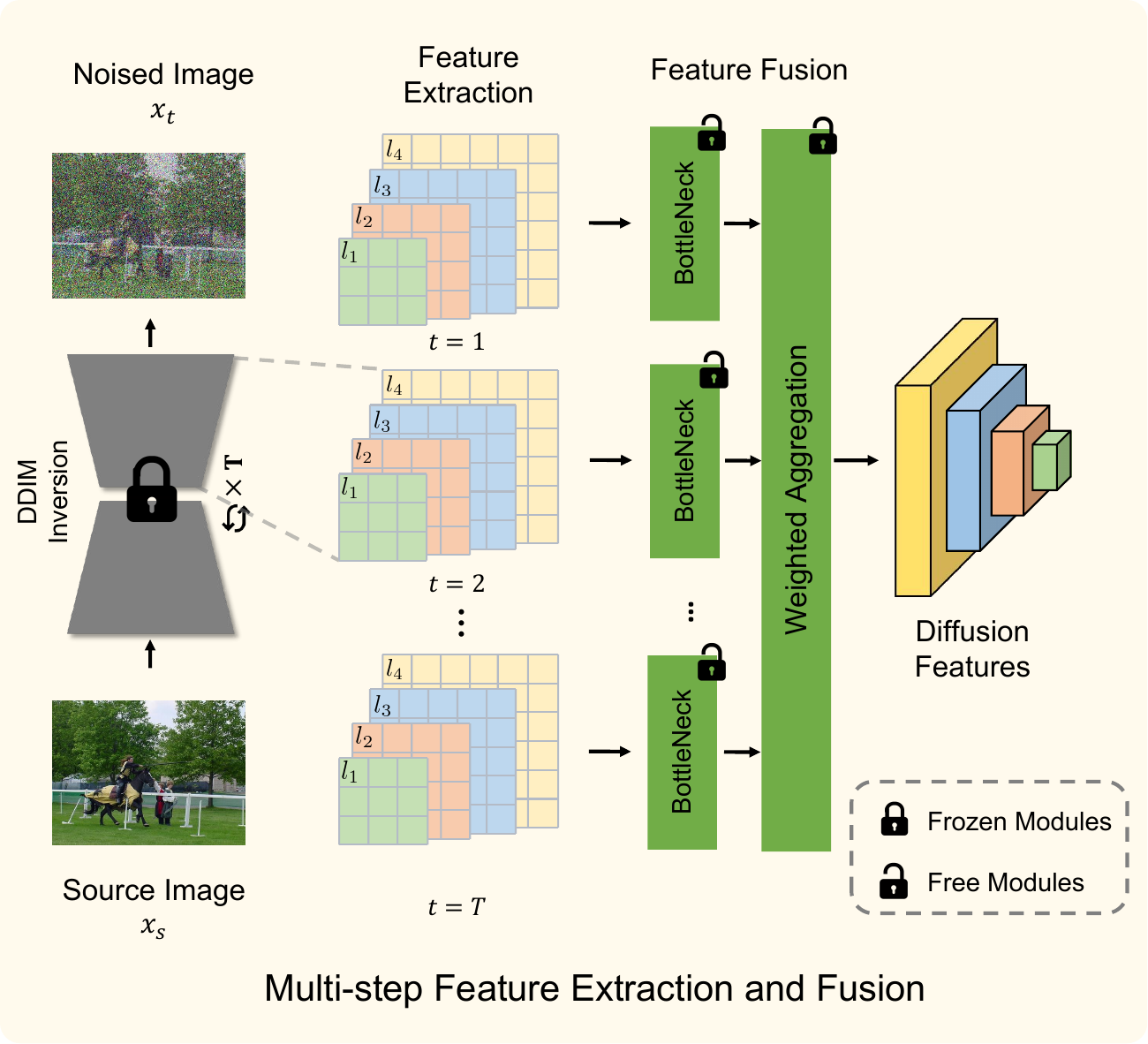}
    \caption{Overview of multi-timestep feature extraction and fusion. Multi-scale features are extracted from a frozen diffusion model at different timesteps, then processed through trainable bottleneck structures and weighted aggregation module to obtain the final hierarchical features.}
    \label{fig:diff backbone}
\end{figure}

\noindent\textbf{Feature Extraction via Diffusion:} The inherent characteristics of diffusion models' intermediate representations make them particularly suitable for domain-invariant feature learning. During the denoising process, the noise predictor $\mathcal{F}_{\theta}$ accumulates multi-scale semantic information by modeling data variations at different noise levels. 

To leverage these properties, we extract and aggregate intermediate features from a sequence of time steps during the forward diffusion process. Specifically, given a source image $\mathbf{x}_s^i$, we progressively add noise following $\mathbf{x}_s^i\rightarrow \mathbf{x}_1\rightarrow \cdots \rightarrow \mathbf{x}_t$, where $t \in \{1,2,\cdots,T\}$. At each time step $t$, we extract features from the four upsampling stages of $\mathcal{F}_{\theta}$, denoted as $\mathbf{s}_t \in \mathbb{R}^{C_{l,k} \times H_{l,k} \times W_{l,k}}$, where for each layer $l \in \{1,2,3,4\}$, we extract three intermediate features ($k=1,2,3$) from the middle of the layer. The feature dimensions are defined as $C_{l,k}$, $H_{l,k}$, and $W_{l,k}$ based on the corresponding layer architecture. This process captures the representation of the transition $\mathbf{x}_{t-1}\rightarrow\mathbf{x}_t$ and generates a multi-timestep feature sequence $\{\mathbf{s}_t\}_{t=1}^T$ for each input image.

\noindent\textbf{Multi-timestep Feature Fusion:} For each timestep, features first go through individual bottleneck structures to align dimensions. A weighted aggregation module then combines features across timesteps with learnable weights as shown in Fig.~\ref{fig:diff backbone}. The fused features form a feature pyramid containing four levels with increasing channel dimensions $C_l \in \{C_1, C_2, C_3, C_4\}$ where $C_l = 256 \times 2^{l-1}$, and corresponding spatial resolutions $H_l \times W_l$ where $H_l = H/2^{l+1}$ and $W_l = W/2^{l+1}$ for $l \in \{1,2,3,4\}$, with $H$ and $W$ being the height and width of input image respectively.

\label{sec:diffusion guided_1}

\subsection{Feature level imitation and alignment}
\label{sec:feature level}
We construct our detector based on the features extracted from the diffusion model as described above. Specifically, we employ Faster R-CNN \cite{faster-rcnn} with all default parameters unchanged and train it exclusively on the source domain to obtain $\mathcal{F}_{\textnormal{diff}}$. 

\noindent\textbf{Motivation:} Common detectors $\mathcal{F}_{\textnormal{comm}}$ tend to overfit source domain data, limiting their generalization to unseen domains. To address this, we propose leveraging knowledge from diffusion-based detector $\mathcal{F}_{\textnormal{diff.}}$ through a two-level alignment approach. Through aligning both feature distributions and object predictions, we expect $\mathcal{F}_{\textnormal{comm}}$ to learn more domain-invariant representations while preserving its detection capability.

\noindent\textbf{Feature alignment:}
\label{sec:diffusion guided_1}
Due to the inherent feature distribution differences between diffusion-based and standard detectors, we adopt PKD~\cite{pkd} for feature alignment, which enables robust cross-architecture knowledge transfer through correlation-based matching.

We extract FPN features $\mathcal{M}_{\textnormal{comm}}^l$ and $\mathcal{M}_{\textnormal{diff}}^l$ from the ResNet~\cite{resnet} of $\mathcal{F}_{\textnormal{comm}}$ and the multi-timestep diffusion feature extraction network in $\mathcal{F}_{\textnormal{diff}}$ respectively. Both features have dimensions $\mathbb{R}^{B \times C \times H_l \times W_l}$ at pyramid level $l$, where $B$ is the batch size, $C$ is the number of channels, and $H_l \times W_l$ represents the spatial dimensions. With $\hat{\mathcal{M}}$ denoting normalized features, the alignment loss is defined as:

\begin{equation}
\mathcal{L}_{\textnormal{align}} = \sum_{l=1}^L \frac{1}{N_l} \|\hat{\mathcal{M}}_{\textnormal{comm}}^l - \hat{\mathcal{M}}_{\textnormal{diff}}^l\|_2^2
\end{equation}

\noindent\textbf{Cross feature adaptation:}
To address the potential instability of direct feature alignment between heterogeneous models, we propose to feed $\mathcal{M}_{\textnormal{diff}}$ into $\mathcal{F}_{\textnormal{comm}}$'s detection heads, enabling stable adaptation while preserving the original detection pipeline. The cross feature loss is defined as:
\begin{equation}
    \mathcal{L}_{\textnormal{cross}} = \mathcal{L}_{\textnormal{comm}}^{\textnormal{rpn}}(\mathcal{M}_{\textnormal{diff}}; \theta_{\textnormal{comm}}) + \mathcal{L}_{\textnormal{comm}}^{\textnormal{roi}}(\mathcal{M}_{\textnormal{diff}}; \theta_{\textnormal{comm}})
\end{equation}
where $\mathcal{L}_{\textnormal{comm}}^{\textnormal{rpn}}$ combines objectness classification and box regression losses for RPN, and $\mathcal{L}_{\textnormal{comm}}^{\textnormal{roi}}$ combines classification and box regression losses for ROI head, following the standard Faster R-CNN detection losses.

\subsection{Domain-invariant Object-level Knowledge Transfer}
\label{sec:object level}
\label{sec:diffusion guided_2}

Beyond feature-level alignment, we aim to enhance $\mathcal{F}_{\textnormal{comm}}$'s object detection capability in unseen domains by transferring task-relevant knowledge from $\mathcal{F}_{\textnormal{diff}}$. However, this presents two challenges: (1) target domain data is unavailable during training under the DG setting, and (2) traditional knowledge distillation methods are not directly applicable due to the heterogeneous architectures between $\mathcal{F}_{\textnormal{comm}}$ and $\mathcal{F}_{\textnormal{diff}}$.

\noindent\textbf{Shared RoI Feature Propagation:}
Inspired by CrossKD~\cite{crosskd}, we propose to align object-level predictions through shared region proposals. We first generate candidate regions $\mathcal{R}_{\textnormal{roi}} \in \mathbb{R}^{N \times 4}$ using the RPN of $\mathcal{F}_{\textnormal{comm}}$, where $N$ is the number of proposals. These regions are used to pool features from both detectors, yielding fixed-size features $\mathcal{M}^{\textnormal{roi}}_{\textnormal{comm}}$ and $\mathcal{M}^{\textnormal{roi}}_{\textnormal{diff}} \in \mathbb{R}^{N \times d}$, where $d$ is the feature dimension. The spatially aligned features are then fed into diffusion detector's branches:

\begin{equation}
    \begin{aligned}
        \mathbf{P}_{\textnormal{cat}} &= \mathcal{F}_{\textnormal{diff}}^{\textnormal{cls}}(\mathcal{M}^{\textnormal{roi}}_{\textnormal{diff}}), \quad \mathbf{P}_{\textnormal{bbox}} = \mathcal{F}_{\textnormal{diff}}^{\textnormal{reg}}(\mathcal{M}^{\textnormal{roi}}_{\textnormal{diff}}) \\
        \mathbf{Q}_{\textnormal{cat}} &= \mathcal{F}_{\textnormal{diff}}^{\textnormal{cls}}(\mathcal{M}^{\textnormal{roi}}_{\textnormal{comm}}), \quad \mathbf{Q}_{\textnormal{bbox}} = \mathcal{F}_{\textnormal{diff}}^{\textnormal{reg}}(\mathcal{M}^{\textnormal{roi}}_{\textnormal{comm}})
    \end{aligned}
\end{equation}

\noindent where $\mathbf{P}_{\textnormal{cat}}, \mathbf{Q}_{\textnormal{cat}} \in \mathbb{R}^{N \times (C+1)}$ denote the class logits for $C$ object categories plus background, and $\mathbf{P}_{\textnormal{bbox}}, \mathbf{Q}_{\textnormal{bbox}} \in \mathbb{R}^{N \times 4}$ represent the predicted box coordinates.

\noindent\textbf{Classification Knowledge Transfer:}
For classification knowledge transfer, we use KL divergence with temperature scaling:

\begin{equation}
    \mathcal{L}_{\textnormal{cls}} = \frac{1}{N} \sum_{i=1}^N \tau^2 D_{KL}(\mathbf{Q}_{\textnormal{cat}}^{i} \| \mathbf{P}_{\textnormal{cat}}^{i})
\end{equation}

\noindent where $\mathbf{Q}_{\textnormal{cat}}^{i}, \mathbf{P}_{\textnormal{cat}}^{i}$ are temperature-scaled softmax outputs for the $i$-th proposal, and $\tau$ is the temperature parameter.

\noindent\textbf{Regression Knowledge Transfer:}
For box regression, we use L1 loss:

\begin{equation}
\mathcal{L}_{\textnormal{reg}} = \frac{1}{N} \sum_{i=1}^N |\mathbf{Q}_{\textnormal{bbox}}^{i} - \mathbf{P}_{\textnormal{bbox}}^{i}|_1
\end{equation}

Through this object-level knowledge transfer mechanism, $\mathcal{F}_{\textnormal{comm}}$ learns domain-invariant detection capabilities from $\mathcal{F}_{\textnormal{diff}}$ via ROI-level alignment. By sharing the same detection head while processing features from different sources, we encourage $\mathcal{F}_{\textnormal{comm}}$ to extract domain-agnostic object features that can generalize well across domains.

\subsection{Joint Optimization Objective}
\label{sec:objective}
The overall objective combines supervised detection learning with feature-level and object-level alignments:

\begin{equation}
    \label{eq:full loss}
    \begin{aligned}
    \mathcal{L}_{\text{total}} = & \underbrace{\mathcal{L}_{\textnormal{det}}(\mathcal{F}_{\textnormal{comm}}(\mathbf{x}_s), \mathbf{y}_s)}_{\text{supervised learning on source domain}} + \\
    & \lambda_{\text{feature}} \underbrace{(\mathcal{L}_{\text{align}} + \mathcal{L}_{\text{cross}})}_{\text{feature-level alignment}} + \lambda_{\text{object}} \underbrace{(\mathcal{L}_{\text{cls}} + \mathcal{L}_{\text{reg}})}_{\text{object-level alignment}}
    \end{aligned}
\end{equation}

\noindent where $\mathcal{L}_{\textnormal{det}}$ is the detection loss on source domain data $(\mathbf{x}_s, \mathbf{y}_s)$, with $\lambda_{\text{feature}}$ and $\lambda_{\text{object}}$ being the weights for feature-level and object-level alignment respectively.

\section{Experiments}
\label{sec:experiments}

\subsection{DG Detection Benchmarks}
\label{sec:benchmarks}
\noindent\textbf{Cross Camera.} Train on Cityscapes~\cite{cityscapes} (2,975 training images from 50 cities) and test on BDD100K~\cite{bdd100k} day-clear split with 7 shared categories following SWDA~\cite{saito2019swda}, evaluating generalization across diverse urban scenes.

\noindent\textbf{Adverse Weather.} Train on Cityscapes~\cite{cityscapes} and test on FoggyCityscapes~\cite{foggy} and RainyCityscapes~\cite{rainy} (synthesized by adding fog and rain effects), using the challenging 0.02 split setting for FoggyCityscapes to evaluate robustness under degraded visibility conditions.

\noindent\textbf{Synthetic to Real.} Train on Sim10K~\cite{sim10k} (10K synthetic driving scenes rendered by GTA-V) and test on Cityscapes~\cite{cityscapes} and BDD100K~\cite{bdd100k} for the \textit{car} category, examining synthetic-to-real transfer capability.

\noindent\textbf{Real to Artistic.} Train on VOC~\cite{voc} (16,551 real-world images from 2007 and 2012) and test on Clipart~\cite{clipart_comic_watercolor} (1K images, 20 categories), Comic~\cite{clipart_comic_watercolor} (2K images, 6 categories), and Watercolor~\cite{clipart_comic_watercolor} (2K images, 6 categories) following~\cite{AT}.

\noindent\textbf{Diverse Weather benchmark.} Train on Daytime-Sunny (26,518 images) and test on four challenging conditions: Night-Sunny (26,158 images), Night-Rainy (2,494 images), Dusk-Rainy (3,501 images), and Daytime-Foggy (3,775 images) following~\cite{cdsd}, evaluating robustness across diverse weather and lighting scenarios. We follow settings from OADG~\cite{oamix} for comparison.

\noindent\textbf{Corruption benchmark.} A comprehensive test-only benchmark~\cite{cityscapes-c} with 15 different corruption types at 5 severity levels for Cityscapes~\cite{cityscapes}, spanning noise, blur, weather, and digital perturbations to evaluate model robustness systematically. We follow settings from OADG~\cite{oamix} for comparison.

\subsection{Implementation Details}
\noindent\textbf{Training settings:} We adopt Faster R-CNN~\cite{faster-rcnn} with ResNet101~\cite{resnet} backbone pretrained on ImageNet~\cite{imagenet} as baseline detector. Models are trained for 20K iterations with batch size 16, learning rate 0.02 and SGD optimizer. We use EMA updated model for stable training. Other settings follow MMDetection defaults~\cite{mmdetection}. 

\label{metrics}
\noindent\textbf{Evaluation metrics:} We report $\text{AP}_{50}$ for individual categories and mAP across categories. For Corruption benchmark~\cite{cityscapes-c}, we additionally report mPC (average $\text{AP}_{50:95}$ across 15 corruptions with 5 levels) and rPC (ratio between mPC and clean performance).

\noindent\textbf{Domain augmentation:} Following~\cite{AT,ddt}, we employ \textit{Strong Augmentation} including both color transformations (color jittering, contrast, equalization, sharpness) and spatial transformations (rotation, shear, translation). Additionally, we design domain-level augmentation strategies by applying \textit{FDA}~\cite{fda}, \textit{Histogram Matching}, and \textit{Pixel Distribution Matching} between source domain images to generate diverse training samples.

\noindent\textbf{Hyper-parameters:} We set diffusion steps $T=5$ and max-timestep as 500 for artistic benchmarks and 100 for other benchmarks as described in Sec.~\ref{sec:diffusion detector}. The loss weights are set as $\lambda_{\text{feature}}=0.5$ and $\lambda_{\text{object}}=1$ in Eq.~\ref{eq:full loss}.

\begin{table}[t]
    \centering
    \caption{Cross Camera DG and DA Results (\%) on BDD100K.}
    \label{tab:bdd100k}
    \setlength{\tabcolsep}{4pt}
    \resizebox{\columnwidth}{!}{%
    \begin{tabular}{l|ccccccc>{\columncolor{gray!20}}c}
    \toprule
    \textbf{Methods} & \textbf{Bike} & \textbf{Bus} & \textbf{Car} & \textbf{Motor} & \textbf{Psn.} & \textbf{Rider} & \textbf{Truck} & \textbf{mAP} \\
    \midrule
    \multicolumn{9}{c}{\textit{\textbf{DG methods} (without target data)}} \\
    CDSD~\cite{cdsd} \textit{\scriptsize{(\textcolor{darkgreen}{CVPR'22})}} & 22.9 & 20.5 & 33.8 & 14.7 & 18.5 & 23.6 & 18.2 & 21.7 \\
    SHADE~\cite{shade} \textit{\scriptsize{(\textcolor{darkgreen}{ECCV'22})}} & 25.1 & 19.0 & 36.8 & 18.4 & 24.1 & 24.9 & 19.8 & 24.0 \\
    SRCD~\cite{srcd} \textit{\scriptsize{(\textcolor{darkgreen}{TNNLS'24})}} & 24.8 & 21.5 & 38.7 & 19.0 & 25.7 & 28.4 & 23.1 & 25.9 \\
    MAD~\cite{mad} \textit{\scriptsize{(\textcolor{darkgreen}{CVPR'23})}} & - & - & - & - & - & - & - & 28.0 \\
    \midrule
    \multicolumn{9}{c}{\textit{\textbf{DA methods} (with unlabeled target data)}} \\ 
    TDD~\cite{tdd} \textit{\scriptsize{(\textcolor{darkgreen}{CVPR'22})}} & 28.8 & 25.5 & 53.9 & 24.5 & 39.6 & 38.9 & 24.1 & 33.6 \\
    PT~\cite{pt} \textit{\scriptsize{(\textcolor{darkgreen}{ICML'22})}}& 28.8 & \textbf{33.8} & 52.7 & 23.0 & 40.5 & 39.9 & 25.8 & 34.9 \\
    SIGMA~\cite{li2022sigma} \textit{\scriptsize{(\textcolor{darkgreen}{CVPR'22})}}& 26.3 & 23.6 & 64.1 & 17.9 & 46.9 & 29.6 & 20.2 & 32.7 \\
    SIGMA++~\cite{li2023sigma++} \textit{\scriptsize{(\textcolor{darkgreen}{TPAMI'23})}}& 27.1 & 26.3 & 65.6 & 17.8 & 47.5 & 30.4 & 21.1 & 33.7 \\
    NSA~\cite{nsa} \textit{\scriptsize{(\textcolor{darkgreen}{ICCV'23})}}& - & - & - & - & - & - & - & 35.5 \\
    HT~\cite{deng2023HT} \textit{\scriptsize{(\textcolor{darkgreen}{CVPR'23})}}& 38.0 & 30.6 & 63.5 & 28.2 & 53.4 & 40.4 & 27.4 & 40.2 \\
    \midrule
    \multicolumn{9}{c}{\textit{\textbf{Ours (DG settings)}}} \\
    \rowcolor{lightgreen} \textbf{Diff. Detector} \footnotesize{(SD-1.5)} & \textbf{38.9} & 31.0 & 71.5 & \underline{37.6} & \underline{61.5} & \textbf{47.0} & \textbf{38.5} & \textbf{46.6} \\
    \textbf{Diff. Detector} \footnotesize{(SD-2.1)} & 38.0 & \underline{33.6} & 69.9 & 36.6 & \textbf{62.1} & 46.3 & 34.2 & 45.8 \\ 
    \textbf{Diff. Guided} \footnotesize{(SD-1.5)} & 38.4 & 33.4 & \textbf{72.0} & \textbf{38.3} & 60.3 & \underline{47.0} & 35.0 & \underline{46.3}\footnotesize{\textcolor{red}{+20.9}} \\
    \textbf{Diff. Guided} \footnotesize{(SD-2.1)} & \underline{38.5} & 32.6 & \underline{71.8} & 37.5 & 60.2 & 46.7 & \underline{35.3} & 46.1\footnotesize{\textcolor{red}{+20.7}} \\
    \bottomrule
    \end{tabular}%
    }
\end{table}

\begin{table}[t]
    \centering
    \caption{Adverse Weather DG and DA Results (\%) on FoggyCityscapes.}
    \label{tab:FoggyCityscapess}
    \setlength{\tabcolsep}{3pt}  
    \resizebox{\columnwidth}{!}{%
    \begin{tabular}{l|cccccccc>{\columncolor{gray!20}}c}
    \toprule
    \textbf{Methods} & \textbf{Bus} & \textbf{Bike} & \textbf{Car} & \textbf{Motor} & \textbf{Psn.} & \textbf{Rider} & \textbf{Train} & \textbf{Truck} & \textbf{mAP} \\
    \midrule
    \multicolumn{10}{c}{\textit{\textbf{DG methods}}} \\
    FACT~\cite{fact} \textit{\scriptsize{(\textcolor{darkgreen}{CVPR'21})}} & 27.7 & 31.3 & 35.9 & 23.3 & 26.2 & 41.2 & 3.0 & 13.6 & 25.3 \\
    FSDR~\cite{fsdr} \textit{\scriptsize{(\textcolor{darkgreen}{CVPR'22})}} & 36.6 & 34.1 & 43.3 & 27.1 & 31.2 & 44.4 & 11.9 & 19.3 & 31.0 \\
    MAD~\cite{mad} \textit{\scriptsize{(\textcolor{darkgreen}{CVPR'23})}} & 44.0 & 40.1 & 45.0 & 30.3 & 34.2 & 47.4 & 42.4 & 25.6 & 38.6 \\
    \midrule
    \multicolumn{10}{c}{\textit{\textbf{DA methods}}} \\
    MGA~\cite{zhou2022mga} \textit{\scriptsize{(\textcolor{darkgreen}{CVPR'22})}} & 53.2 & 36.9 & 61.5 & 27.9 & 43.1 & 47.3 & 50.3 & 30.2 & 43.8 \\
    MTTrans~\cite{MTTrans} \textit{\scriptsize{(\textcolor{darkgreen}{CVPR'22})}} & 45.9 & 46.5 & 65.2 & 32.6 & 47.7 & 49.9 & 33.8 & 25.8 & 43.4 \\
    OADA~\cite{oada} \textit{\scriptsize{(\textcolor{darkgreen}{CVPR'22})}} & 48.5 & 39.8 & 62.9 & 34.3 & 47.8 & 46.5 & 50.9 & 32.1 & 45.4 \\
    MIC~\cite{mic} \textit{\scriptsize{(\textcolor{darkgreen}{CVPR'23})}}& 52.4 & 47.5 & 67.0 & 40.6 & 50.9 & 55.3 & 33.7 & 33.9 & 47.6 \\
    SIGMA++~\cite{li2023sigma++} \textit{\scriptsize{(\textcolor{darkgreen}{TPAMI'23})}}  & 52.2 & 39.9 & 61.0 & 34.8 & 46.4 & 45.1 & 44.6 & 32.1 & 44.5 \\
    CIGAR~\cite{CIGAR} \textit{\scriptsize{(\textcolor{darkgreen}{CVPR'23})}} & \underline{56.6} & 41.3 & 62.1 & 33.7 & 46.1 & 47.3 & 44.3 & 27.8 & 44.9 \\
    CMT~\cite{cao2023cmt} \textit{\scriptsize{(\textcolor{darkgreen}{CVPR'23})}} & \textbf{66.0} & 51.2 & 63.7 & 41.4 & 45.9 & 55.7 & 38.8 & \textbf{39.6} & 50.3 \\
    HT~\cite{deng2023HT} \textit{\scriptsize{(\textcolor{darkgreen}{CVPR'23})}} & 55.9 & 50.3 & \textbf{67.5} & 40.1 & \textbf{52.1} & 55.8 & 49.1 & 32.7 & 50.4 \\
    \midrule
    \multicolumn{10}{c}{\textit{\textbf{Ours (DG settings)}}} \\
    \textbf{Diff. Detector} \footnotesize{(SD-1.5)} & 56.2 & 50.4 & 66.7 & 39.9 & 50.2 & \underline{59.5} & 39.9 & \underline{38.0} & 50.1 \\
    \textbf{Diff. Detector} \footnotesize{(SD-2.1)} & 55.5 & 49.6 & 67.0 & 40.4 & 50.4 & 58.2 & 29.2 & 36.4 & 48.3 \\
    \rowcolor{lightgreen}
    \textbf{Diff. Guided} \footnotesize{(SD-1.5)} & 53.8 & \textbf{54.2} & \textbf{67.5} & \textbf{45.6} & \textbf{52.1} & \textbf{60.8} & \textbf{53.9} & 32.4 & \textbf{52.5}\footnotesize{\textcolor{red}{+21.8}} \\
    \textbf{Diff. Guided} \footnotesize{(SD-2.1)} & 55.1 & \underline{53.9} & 67.0 & \underline{43.4} & \underline{51.9} & \underline{59.5} & 42.2 & 34.8 & \underline{51.0}\footnotesize{\textcolor{red}{+20.3}} \\
    \bottomrule
    \end{tabular}%
    }
\end{table}

\begin{table}[t]
    \begin{minipage}[t]{0.20\textwidth}
        \centering
        \setlength{\tabcolsep}{1pt}
        \caption{Adverse Weather DG and DA Results (\%) on RainyCityscapes.}
        \label{tab:RainyCityscapes}
        \resizebox{\textwidth}{!}{%
        \begin{tabular}{l|c>{\columncolor{gray!20}}c}
        \toprule
        \textbf{Methods} & \textbf{mAP} \\
        \midrule
        \multicolumn{2}{c}{\textit{\textbf{DG methods}}} \\
        FACT~\cite{fact} \textit{\scriptsize{(\textcolor{darkgreen}{CVPR'21})}} & 39.9 \\
        FSDR~\cite{fsdr} \textit{\scriptsize{(\textcolor{darkgreen}{CVPR'22})}}& 42.8 \\
        SCG~\cite{mad} \textit{\scriptsize{(\textcolor{darkgreen}{CVPR'23})}}& 39.1 \\
        MAD~\cite{mad} \textit{\scriptsize{(\textcolor{darkgreen}{CVPR'23})}}& 42.3 \\
        \midrule
        \multicolumn{2}{c}{\textit{\textbf{DA methods}}} \\
        MGA~\cite{zhou2022mga} \textit{\scriptsize{(\textcolor{darkgreen}{CVPR'22})}}& 43.0 \\
        TDD~\cite{tdd} \textit{\scriptsize{(\textcolor{darkgreen}{CVPR'23})}}& 50.3 \\
        CMT~\cite{cao2023cmt} \textit{\scriptsize{(\textcolor{darkgreen}{CVPR'23})}}& 52.1 \\
        SIGMA++~\cite{li2023sigma++} \textit{\scriptsize{(\textcolor{darkgreen}{TPAMI'23})}}& 46.9 \\
        \midrule
        \multicolumn{2}{c}{\textit{\textbf{Ours (DG settings)}}} \\
        \textbf{Diff. Detector} \footnotesize{(SD-1.5)} & \underline{58.2} \\
        \textbf{Diff. Detector} \footnotesize{(SD-2.1)} & 56.1 \\ 
        \textbf{Diff. Guided} \footnotesize{(SD-1.5)} & 57.9\footnotesize{\textcolor{red}{+21.5}} \\
        \rowcolor{lightgreen} \textbf{Diff. Guided} \footnotesize{(SD-2.1)} & \textbf{58.3}\footnotesize{\textcolor{red}{+21.9}} \\
        \bottomrule
        \end{tabular}%
        }
    \end{minipage}
    \hfill
    \begin{minipage}[t]{0.26\textwidth}
        \centering
        \setlength{\tabcolsep}{2pt}
        \caption{Synthetic to Real DG and DA Results (\%) of category \textit{car} on Cityscapes and BDD100K.}
        \label{tab:sim10k}
        \resizebox{\textwidth}{!}{%
        \begin{tabular}{l|cc>{\columncolor{gray!20}}c}
        \toprule
        \textbf{Methods} & \textbf{Cityscapes} & \textbf{BDD100K} \\
        \midrule
        \multicolumn{3}{c}{\textit{\textbf{DG methods}}} \\
        CDSD~\cite{cdsd} \textit{\scriptsize{(\textcolor{darkgreen}{CVPR'22})}} & 35.2 & 27.4 \\
        SHADE~\cite{shade} \textit{\scriptsize{(\textcolor{darkgreen}{CVPR'22})}}& 40.9 & 30.3 \\
        SRCD~\cite{srcd} \textit{\scriptsize{(\textcolor{darkgreen}{TNNLS'24})}} & 43.0 & 31.6 \\
        \midrule
        \multicolumn{3}{c}{\textit{\textbf{DA methods}}} \\
        SWDA~\cite{saito2019swda} \textit{\scriptsize{(\textcolor{darkgreen}{CVPR'19})}} & 40.7 & 42.9 \\
        MTTrans~\cite{MTTrans} \textit{\scriptsize{(\textcolor{darkgreen}{CVPR'22})}} & 57.9 & - \\
        SIGMA~\cite{li2022sigma} \textit{\scriptsize{(\textcolor{darkgreen}{CVPR'22})}} & 53.7 & - \\
        TDD~\cite{tdd} \textit{\scriptsize{(\textcolor{darkgreen}{CVPR'22})}} & 53.4 & - \\
        MGA~\cite{zhou2022mga} \textit{\scriptsize{(\textcolor{darkgreen}{CVPR'22})}} & 54.1 & - \\
        SIGMA++~\cite{li2023sigma++} \textit{\scriptsize{(\textcolor{darkgreen}{TPAMI'23})}} & 53.7 & - \\
        CIGAR~\cite{CIGAR} \textit{\scriptsize{(\textcolor{darkgreen}{CVPR'23})}} & 58.5 & - \\
        NSA~\cite{nsa} \textit{\scriptsize{(\textcolor{darkgreen}{ICCV'23})}} & 56.3 & - \\
        \midrule
        \multicolumn{3}{c}{\textit{\textbf{Ours (DG settings)}}} \\
        \rowcolor{lightgreen} \textbf{Diff. Detector} \footnotesize{(SD-1.5)} & \underline{62.8} & \textbf{64.4} \\
        \rowcolor{lightgreen} \textbf{Diff. Detector} \footnotesize{(SD-2.1)} & \textbf{64.5} & \underline{64.1} \\ 
        \textbf{Diff. Guided} \footnotesize{(SD-1.5)} & 59.7\footnotesize{\textcolor{red}{+22.3}} & 58.2\footnotesize{\textcolor{red}{+30.0}} \\
        \textbf{Diff. Guided} \footnotesize{(SD-2.1)} & 57.3\footnotesize{\textcolor{red}{+19.9}} & 54.5\footnotesize{\textcolor{red}{+26.3}} \\
        \bottomrule
        \end{tabular}%
        }
    \end{minipage}
\end{table}

\begin{table}[t]
    \centering
    \caption{Generalization detection Results (\%) on Diverse Weather benchmark. \textbf{DF}: Daytime-Foggy, \textbf{DR}: Dusk-Rainy, \textbf{NR}: Night-Rainy, \textbf{NS}: Night-Sunny, as described in Sec.~\ref{sec:benchmarks}. 
    }
    \label{tab:dwd}
    \setlength{\tabcolsep}{10pt}
    \resizebox{1\columnwidth}{!}{%
    \begin{tabular}{l|cccc|>{\columncolor{gray!20}}c}
    \toprule
    \textbf{Methods} & \textbf{DF} & \textbf{DR} & \textbf{NR} & \textbf{NS} & \textbf{Average} \\ \midrule
    CDSD~\cite{cdsd} \textit{\scriptsize{(\textcolor{darkgreen}{CVPR'22})}} & 33.5 & 28.2 & 16.6 & 36.6 & 28.7 \\
    SHADE~\cite{shade} \textit{\scriptsize{(\textcolor{darkgreen}{CVPR'22})}} & 33.4 & 29.5 & 16.8 & 33.9 & 28.4 \\
    CLIPGap~\cite{clip_gap} \textit{\scriptsize{(\textcolor{darkgreen}{CVPR'23})}} & 32.0 & 26.0 & 12.4 & 34.4 & 26.2 \\
    SRCD~\cite{srcd} \textit{\scriptsize{(\textcolor{darkgreen}{TNNLS'24})}} & 35.9 & 28.8 & 17.0 & 36.7 & 29.6 \\
    G-NAS~\cite{gnas} \textit{\scriptsize{(\textcolor{darkgreen}{AAAI'24})}} & 36.4 & 35.1 & 17.4 & 45.0 & 33.5 \\
    OA-DG~\cite{oamix} \textit{\scriptsize{(\textcolor{darkgreen}{AAAI'24})}} & 38.3 & 33.9 & 16.8 & 38.0 & 31.8 \\
    DivAlign~\cite{Diversification} \textit{\scriptsize{(\textcolor{darkgreen}{CVPR'24})}} & 37.2 & 38.1 & \underline{24.1} & 42.5 & 35.5 \\
    UFR~\cite{ufr} \textit{\scriptsize{(\textcolor{darkgreen}{CVPR'24})}} & 39.6 & 33.2 & 19.2 & 40.8 & 33.2 \\ \midrule \midrule
    \rowcolor{lightgreen}
    \textbf{Diff. Detector} \footnotesize{(SD-1.5)} & 43.3 & \textbf{42.5} & \textbf{27.8} & 47.0 & \textbf{40.2} \\
    \textbf{Diff. Detector} \footnotesize{(SD-2.1)} & 44.6 & \underline{41.6} & 23.2 & 46.4 & \underline{39.0} \\
    \midrule
    \textbf{Diff. Guided} \footnotesize{(SD-1.5)} & \textbf{44.7} & 37.4 & 21.7 & \underline{48.7} & 38.1\footnotesize{\textcolor{red}{+13.9}} \\
    \textbf{Diff. Guided} \footnotesize{(SD-2.1)} & \underline{44.7} & 37.1 & 20.0 & \textbf{49.3} & 37.8\footnotesize{\textcolor{red}{+13.6}} \\ \bottomrule
    \end{tabular}%
    }
\end{table}

\begin{table}[t]
    \centering
    \caption{Real to Artistic DG and DA Results (\%) on Clipart, Comic, Watercolor.} 
    \label{tab:artistic}
    \setlength{\tabcolsep}{13pt}
    \resizebox{\columnwidth}{!}{%
    \begin{tabular}{l|ccc}
    \toprule
    \textbf{Methods} & \textbf{Clipart} & \textbf{Comic} & \textbf{Watercolor} \\ \midrule
    \multicolumn{4}{c}{\textit{\textbf{DG methods}}} \\
    Div. \textit{\scriptsize{(\textcolor{darkgreen}{CVPR'24})}} & 33.7 & 25.5 & 52.5 \\
    DivAlign \textit{\scriptsize{(\textcolor{darkgreen}{CVPR'24})}} & 38.9 & 33.2 & 57.4 \\ \midrule
    \multicolumn{4}{c}{\textit{\textbf{DA methods}}} \\
    SWDA \textit{\scriptsize{(\textcolor{darkgreen}{CVPR'19})}} & -- & 29.4 & 53.3 \\
    MCRA \textit{\scriptsize{(\textcolor{darkgreen}{ECCV'20})}} & -- & 33.5 & 56.0 \\
    I3Net \textit{\scriptsize{(\textcolor{darkgreen}{CVPR'21})}} & -- & 30.1 & 51.5 \\
    DBGL \textit{\scriptsize{(\textcolor{darkgreen}{ICCV'21})}} & -- & 29.7 & 53.8 \\
    AT \textit{\scriptsize{(\textcolor{darkgreen}{CVPR'22})}} & 49.3 & -- & 59.9 \\
    D-ADAPT \textit{\scriptsize{(\textcolor{darkgreen}{ICLR'22})}} & 49.0 & 40.5 & -- \\
    TIA \textit{\scriptsize{(\textcolor{darkgreen}{CVPR'22})}} & 46.3 & -- & -- \\
    LODS \textit{\scriptsize{(\textcolor{darkgreen}{CVPR'22})}} & 45.2 & -- & 58.2 \\
    CIGAR \textit{\scriptsize{(\textcolor{darkgreen}{CVPR'23})}} & 46.2 & -- & -- \\
    CMT \textit{\scriptsize{(\textcolor{darkgreen}{CVPR'23})}} & 47.0 & -- & -- \\ 
    \midrule
    \multicolumn{4}{c}{\textit{\textbf{Ours (DG settings)}}} \\
    \rowcolor{yellow!15}
    \textbf{Diff. Detector} \footnotesize{(SD-1.5)} & \textbf{58.3} & \textbf{51.9} & \textbf{68.4} \\
    \textbf{Diff. Detector} \footnotesize{(SD-2.1)} & \underline{51.7} & \underline{46.6} & \underline{62.1} \\
    \textbf{Diff. Guided} \footnotesize{(SD-1.5)} & 40.8\footnotesize{\textcolor{red}{+13.6}} & 29.7\footnotesize{\textcolor{red}{+11.6}} & 54.2\footnotesize{\textcolor{red}{+12.7}} \\
    \textbf{Diff. Guided} \footnotesize{(SD-2.1)} & 32.7\footnotesize{\textcolor{red}{+5.5}} & 24.9\footnotesize{\textcolor{red}{+6.8}} & 50.6\footnotesize{\textcolor{red}{+9.1}} \\ \bottomrule
    \end{tabular}%
    }
\end{table}

\begin{table*}[t]
    \centering
    \caption{Generalization detection Results (\%) on Cityscapes Corruption benchmark. (mPC and rPC are defined in Sec.~\ref{metrics}).}
    \label{tab:cityscapes-c}
    \setlength{\tabcolsep}{3pt}
    \resizebox{\textwidth}{!}{%
    \begin{tabular}{l|c|ccc|cccc|ccc|ccccc|>{\columncolor{gray!20}}c>{\columncolor{gray!20}}c}
    \toprule
    & & \multicolumn{3}{c|}{\textbf{Noise}} & \multicolumn{4}{c|}{\textbf{Blur}} & \multicolumn{3}{c|}{\textbf{Weather}} & \multicolumn{5}{c|}{\textbf{Digital}} & \multicolumn{2}{c}{} \\
    \textbf{Methods} & Clean & Gauss. & Shot & Impulse & Defocus & Glass & Motion & Zoom & Snow & Frost & Fog & Bright & Contrast & Elastic & JPEG & Pixel & \textbf{mPC \textcolor{red}{$\uparrow$}} & \textbf{rPC \textcolor{red}{$\uparrow$}} \\
    \midrule
    FSCE~\cite{fsce} \textit{\scriptsize{(\textcolor{darkgreen}{CVPR'21})}} & 43.1 & 7.4 & 10.2 & 8.2 & 23.3 & 20.3 & 21.5 & \underline{4.8} & 5.6 & 23.6 & 37.1 & 38.0 & 31.9 & \underline{40.0} & 20.4 & 23.2 & 21.0 & 48.7 \\
    OA-Mix~\cite{oamix} \textit{\scriptsize{(\textcolor{darkgreen}{AAAI'24})}} & 42.7 & 7.2 & 9.6 & 7.7 & 22.8 & 18.8 & 21.9 & \textbf{5.4} & 5.2 & 23.6 & 37.3 & 38.7 & 31.9 & \textbf{40.2} & 20.2 & 22.2 & 20.8 & 48.7 \\
    OA-DG~\cite{oamix} \textit{\scriptsize{(\textcolor{darkgreen}{AAAI'24})}}& 43.4 & 8.2 & 10.6 & 8.4 & 24.6 & \underline{20.5} & \underline{22.3} & \underline{4.8} & 6.1 & \textbf{25.0} & 38.4 & \underline{39.7} & 32.8 & \textbf{40.2} & 22.0 & \underline{23.8} & 21.8 & 50.2 \\
    \midrule \midrule
    \rowcolor{lightgreen} \textbf{Diff. Detector} \footnotesize{(SD-1.5)} & 34.7 & \textbf{20.3} & \textbf{23.2} & \textbf{17.2} & \textbf{26.8} & \textbf{21.7} & \textbf{23.7} & 3.4 & \textbf{16.6} & 24.2 & 32.5 & 34.4 & 30.6 & 33.7 & \textbf{29.1} & \textbf{24.4} & \textbf{24.1} & \textbf{69.5} \\
    \textbf{Diff. Detector} \footnotesize{(SD-2.1)} & 34.7 & \underline{18.4} & \underline{20.9} & \underline{15.7} & \underline{26.2} & \underline{20.5} & 21.9 & 4.0 & \underline{14.3} & 22.6 & 31.4 & 33.8 & 29.3 & 32.5 & \underline{27.9} & 21.5 & \underline{22.7} & \underline{65.5} \\ \midrule
    \textbf{Diff. Guided} \footnotesize{(SD-1.5)} & 42.1 & 11.0 & 13.6 & 10.8 & 25.0 & 14.2 & 21.4 & 3.4 & 5.4 & 24.0 & \textbf{39.6} & \textbf{40.3} & \textbf{36.3} & 39.2 & 18.9 & 16.0 & 21.3\footnotesize{\textcolor{red}{+5.7}} & 50.5\footnotesize{\textcolor{red}{+12.3}} \\
    \textbf{Diff. Guided} \footnotesize{(SD-2.1)} & 42.2 & 8.2 & 10.5 & 8.2 & 21.6 & 12.4 & 20.1 & 3.0 & 3.1 & \underline{24.5} & \underline{39.2} & 39.5 & \underline{35.8} & 38.7 & 23.1 & 19.6 & 20.5\footnotesize{\textcolor{red}{+4.9}} & 48.6\footnotesize{\textcolor{red}{+10.4}} \\
    \bottomrule
    \end{tabular}%
    }
\end{table*}

\subsection{Results and Comparisons}
We compare our approach against existing DG methods (target domain unseen) and DA methods (target domain unlabeled). Our results include \textbf{Diff. Detector} trained solely on source domain (Sec.~\ref{sec:diffusion detector}), \textbf{SD-1.5} and \textbf{SD-2.1} using different StableDiffusion~\cite{rombach2022latent} versions, and \textbf{Diff. Guided} which applies our alignment approach to Faster R-CNN~\cite{faster-rcnn} baseline through \textbf{Diff. Detector} as described in Sec.~\ref{sec:diffusion guided_1} and~\ref{sec:diffusion guided_2}.

In all tables, \textbf{bold} and \underline{underline} denote the best and second-best results. \colorbox{lightgreen}{Yellow background} highlights the best average performance. And {\textcolor{red}{+{$x$}}} indicates mAP(\%) gains over baseline.

We conduct extensive experiments across six challenging benchmarks to evaluate our method in Tab.~\ref{tab:bdd100k}, ~\ref{tab:FoggyCityscapess}, ~\ref{tab:RainyCityscapes}, ~\ref{tab:sim10k}, ~\ref{tab:artistic}, ~\ref{tab:dwd} and  ~\ref{tab:cityscapes-c}. Our comprehensive evaluations demonstrate that the diffusion detector consistently achieves SOTA performance in DG settings and even surpasses most domain adaptation methods that require target domain data. Through effective feature and object alignment, our diffusion guidance mechanism successfully enhances detector generalization under moderate domain gaps. However, its improvement becomes more limited when facing extreme domain shifts, particularly in Real to Artistic benchmarks (in Tab.~\ref{tab:artistic}).

\begin{table}[h]
    \centering
    \begin{tabular}{cc}
        \setlength{\tabcolsep}{2pt}
          \begin{minipage}{0.2\textwidth}
            \renewcommand{\arraystretch}{1.25}
            \centering
            \caption{Testing results and inference costs of different diffusion steps $T$.}
            \label{tab:diffusion_steps}
            \resizebox{\textwidth}{!}{
                \begin{tabular}{c|ccc|c}
                    \toprule
                    {\footnotesize T} & 
                    \begin{tabular}[c]{@{}c@{}}{\footnotesize BDD}\\[-0.5ex] {\footnotesize 100K}\end{tabular} & 
                    \begin{tabular}[c]{@{}c@{}}{\footnotesize Cityscapes}\\[-0.5ex] {\footnotesize (car)}\end{tabular} & 
                    {\footnotesize Clipart} & 
                    \begin{tabular}[c]{@{}c@{}}{\footnotesize Inference}\\[-0.5ex] {\footnotesize Time (ms)}\end{tabular} \\ 
                    \midrule
                    1 & 28.6 & 49.8 & 37.4 & 270 \\
                    2 & 34.9 & 54.1 & 48.6 & 404 \\
                    \textbf{\underline{5}} & 46.6 & \textbf{62.8} & 58.3 & 789 \\
                    10 & \textbf{47.1} & 62.6 & \textbf{58.9} & 1,424 \\
                    20 & 45.6 & 61.4 & 57.7 & 2,820 \\
                    \bottomrule
                \end{tabular}
        }
      \end{minipage}
      \begin{minipage}{0.25\textwidth}
        \centering  
        \includegraphics[width=\linewidth]{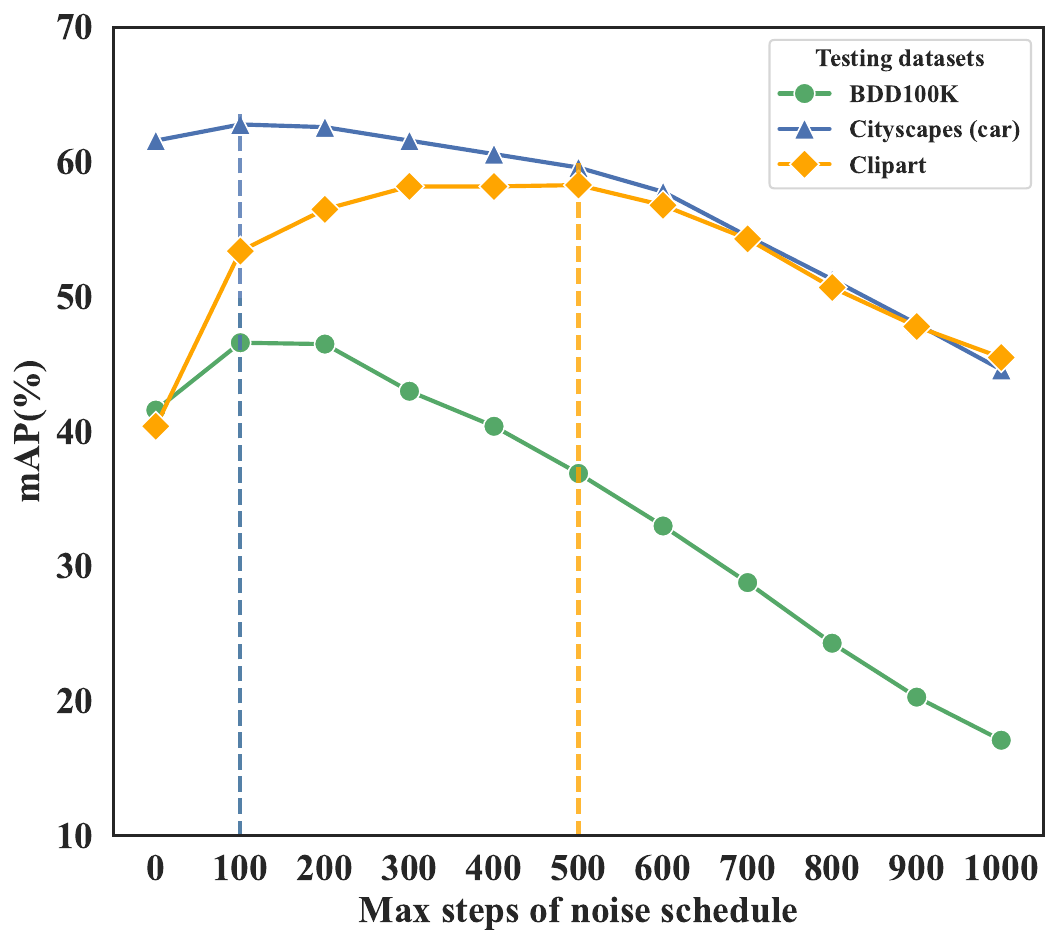} 
        \captionof{figure}{Testing results of different max timesteps.}
        \label{fig:max steps}
      \end{minipage} &
    \end{tabular}
  \end{table}


\begin{table}[h]
    \centering
    \caption{Comparison of Diffusion backbone and stronger models}
    \label{tab:backbone-comparison}
    \setlength{\tabcolsep}{4pt}
    \resizebox{1\columnwidth}{!}{%
    \begin{tabular}{l|ccc|cccc}
    \toprule
    \textbf{Models} & \textbf{Clipart} & \textbf{Comic} & \textbf{Watercolor} & \textbf{DF} & \textbf{DR} & \textbf{NR} & \textbf{NS} \\ \midrule
    ConvNeXt-base~\cite{Convnext} & 43.6 & 26.6 & 55.1 & 39.7 & 39.2 & 23.4 & 45.9 \\
    VIT-base~\cite{vit}& 29.5 & 15.5 & 43.0 & 24.8 & 25.8 & 11.4 & 23.0 \\
    Swin-base~\cite{swin}& 30.2 & 18.0 & 42.6 & 37.2 & 38.9 & 22.6 & 42.2 \\
    MAE (VIT-base)~\cite{mae} & 28.1 & 16.7 & 44.4 & 32.5 & 32.6 & 17.0 & 34.3 \\
    Glip (Swin-tiny)~\cite{glip} & 39.2 & 18.7 & 50.4 & 38.5 & 36.3 & 20.3 & 45.5 \\ \midrule
    \textbf{Diffusion backbone}& \textbf{58.3} & \textbf{51.9} & \textbf{68.4} & \textbf{43.4} & \textbf{42.5} & \textbf{27.8} & \textbf{47.0} \\ 
    \bottomrule
    \end{tabular}%
    }
\end{table}

\section{Ablation Studies}

\begin{figure}[ht]
    \centering
    \begin{minipage}{\columnwidth}
        \centering
        \captionof{table}{Model calibration performance with D-ECE~\cite{dece} metric.}
        \resizebox{\columnwidth}{!}{%
        \setlength{\tabcolsep}{8pt}
        \begin{tabular}{c|ccc}
        \toprule
                         & \multicolumn{3}{c}{\textbf{D-ECE(\%) \textcolor{red}{$\downarrow$}}}       \\
        \textbf{Detector}         & \textbf{BDD100K} & \textbf{Cityscapes (car)} & \textbf{Clipart} \\ \midrule
        FR-R101 Baseline & 10.9    & 20.2             & 10.8    \\
        \textbf{Diff. Detector}   & \textbf{8.5}     & \textbf{8.7}              & \textbf{5.5}     \\ \bottomrule
        \end{tabular}
        }
        \label{tab:dece}
    \end{minipage}
    
    \begin{minipage}{\columnwidth}
        \centering
        \includegraphics[width=0.96\textwidth]{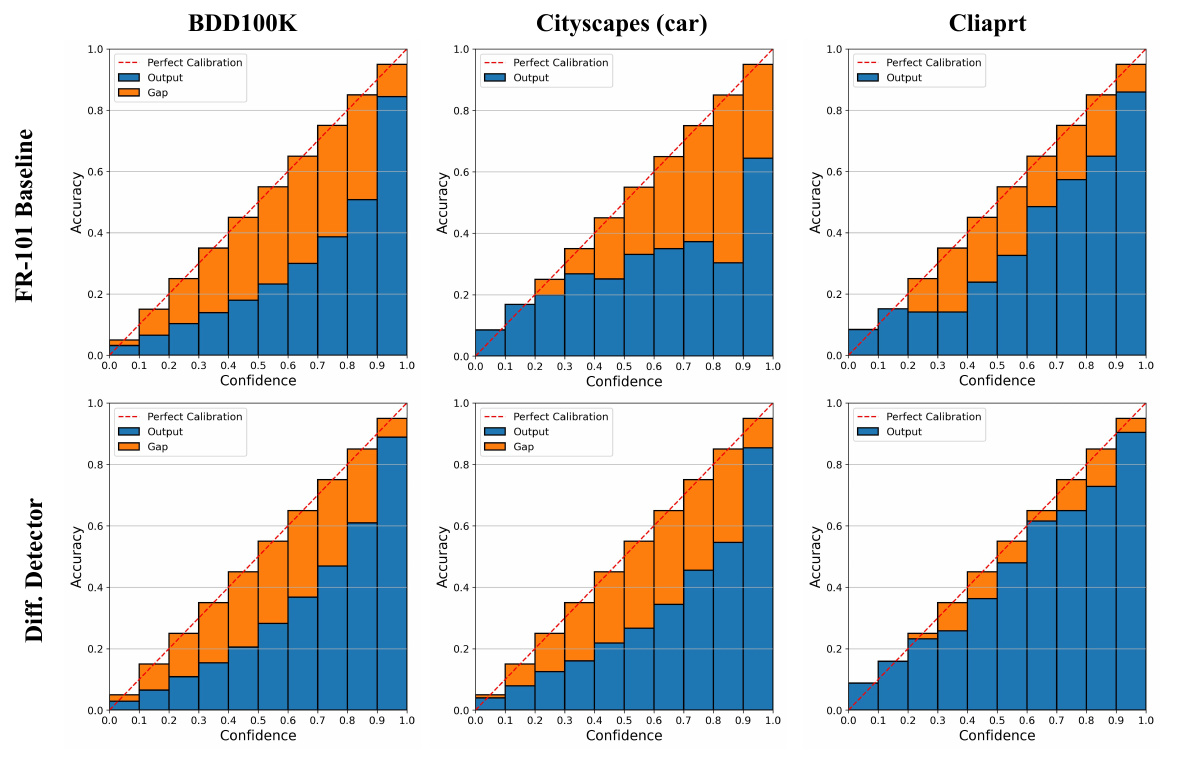}
        \caption{Reliability Diagram on different target domains. Curves closer to the diagonal indicate better performance.}
        \label{fig:dece}
    \end{minipage}
\end{figure}

\label{sec:ablation}
\subsection{Studies on Diffusion Detector}

\noindent\textbf{Comparison with stronger backbones:} We evaluate against recent advanced models~\cite{Convnext,vit,swin,mae,glip} under same settings. Results in Tab.~\ref{tab:backbone-comparison} demonstrate our diffusion backbone's superior generalization through effective domain-invariant feature extraction.

We investigate two key parameters for extracting features from diffusion models: the number of diffusion steps $T$ and max timesteps (Sec.~\ref{sec:diffusion detector}).

\noindent\textbf{Analysis of diffusion steps $T$:} Tab.~\ref{tab:diffusion_steps} shows that larger $T$ values improve performance but increase inference time. We set $T=5$ as default for balancing accuracy and efficiency.

\noindent\textbf{Analysis of max timesteps:} Fig.~\ref{fig:max steps} demonstrates that larger timesteps (e.g., 500) benefit benchmarks with severe domain shifts like Real-to-Artistic, while others perform well at timestep 100.

\noindent\textbf{Insights on generalization:} Our empirical results reveal two key findings: (1) Without noise diffusion (max timestep 0), the model inherits strong transfer capability from large-scale pre-training, similar to GLIP~\cite{glip}; (2) The noise-adding and denoising process enhances generalization, with higher noise levels particularly benefiting larger domain shifts through learning domain-invariant features.

\subsection{Studies on Diffusion Guided Detector}
\noindent\textbf{Analysis of proposed modules:} Tab.~\ref{tab:study} validates our components' effectiveness for domain generalization. Domain augmentation yields limited gains (\textbf{2.7\%}) for the diffusion detector but significant improvement (\textbf{11.0\%}) for baseline, indicating diffusion detector relies more on inherent generalization while trainable-backbone detectors benefit more from diverse training samples.

Feature-level and object-level alignment improve performance by \textbf{3.5\%} and \textbf{6.4\%} respectively, showing effective representation learning under diffusion guidance. The consistent gains (\textbf{6.1\%}) with data augmentation further demonstrate our alignment approach's compatibility with conventional techniques.

\begin{table}[ht]
    \centering
    \renewcommand{\arraystretch}{1.25}
    \caption{Ablation studies of our framework components. \textbf{Settings}: domain augmentation (\textbf{Aug.}), feature-level alignment (\textbf{Fea.}), and object-level alignment (\textbf{Obj.}). Foggy Cityscapes (\textbf{F-C}), Rainy Cityscapes (\textbf{R-C})}
    \label{tab:study}
    \setlength{\tabcolsep}{1pt}
    \resizebox{\columnwidth}{!}{%
    \begin{tabular}{l|ccc|ccc|cccc}
    \toprule
    \textbf{Settings}                 & \textbf{Aug.} & \textbf{Fea.}   & \textbf{Obj.}  & \textbf{BDD}  & \textbf{F-C}  & \textbf{R-C}  & \textbf{DF}   & \textbf{DR}   & \textbf{NR}   & \textbf{NS}   \\ \midrule
    \multirow{2}{*}{\makecell{Diff.\\Detector}}   &               & \multicolumn{2}{c|}{}           & 44.5          & 48.0          & 54.4          & 41.3          & 40.4          & 24.3          & 43.7          \\
                                      & \checkmark    & \multicolumn{2}{c|}{}           & \textbf{46.6}\footnotesize{\textcolor{red}{+2.1}}          & \textbf{50.1}\footnotesize{\textcolor{red}{+2.1}}          & \textbf{58.2}\footnotesize{\textcolor{red}{+3.8}}          & \textbf{43.3}\footnotesize{\textcolor{red}{+2.0}}          & \textbf{42.5}\footnotesize{\textcolor{red}{+2.1}}          & \textbf{27.8}\footnotesize{\textcolor{red}{+3.5}}          & \textbf{47.0}\footnotesize{\textcolor{red}{+3.3}}          \\ \midrule \midrule
    \multirow{2}{*}{\makecell{FR-R101\\baseline}} &               & \multicolumn{2}{c|}{}           & 25.4          & 30.7          & 36.4          & 28.8          & 24.1          & 12.4          & 31.4          \\
                                      & \checkmark    & \multicolumn{2}{c|}{}           & 36.2\footnotesize{\textcolor{red}{+10.8}}          & 47.9\footnotesize{\textcolor{red}{+17.2}}          & 50.1\footnotesize{\textcolor{red}{+13.7}}          & 39.9\footnotesize{\textcolor{red}{+11.1}}          & 34.8\footnotesize{\textcolor{red}{+10.7}}          & 16.4\footnotesize{\textcolor{red}{+4.0}}          & 41.2\footnotesize{\textcolor{red}{+9.8}}          \\ \midrule
    \multirow{4}{*}{\makecell{Diff.\\Guided}}     &               & \checkmark          &                   & 29.8\footnotesize{\textcolor{red}{+4.4}}          & 34.2\footnotesize{\textcolor{red}{+3.5}}          & 41.2\footnotesize{\textcolor{red}{+4.8}}          & 34.2\footnotesize{\textcolor{red}{+5.4}}          & 24.9\footnotesize{\textcolor{red}{+0.8}}          & 14.4\footnotesize{\textcolor{red}{+2.0}}          & 35.2\footnotesize{\textcolor{red}{+3.8}}          \\
                                      &               &                     & \checkmark        & 34.9\footnotesize{\textcolor{red}{+9.5}}          & 38.3\footnotesize{\textcolor{red}{+7.6}}          & 43.1\footnotesize{\textcolor{red}{+6.7}}          & 37.2\footnotesize{\textcolor{red}{+8.4}}          & 26.3\footnotesize{\textcolor{red}{+2.2}}          & 16.4\footnotesize{\textcolor{red}{+4.0}}          & 38.1\footnotesize{\textcolor{red}{+6.7}}          \\
                                      &               & \checkmark          & \checkmark        & 38.1\footnotesize{\textcolor{red}{+12.7}}          & 40.4\footnotesize{\textcolor{red}{+9.7}}          & 45.9\footnotesize{\textcolor{red}{+9.5}}          & 38.6\footnotesize{\textcolor{red}{+9.8}}          & 28.4\footnotesize{\textcolor{red}{+4.3}}          & 16.8\footnotesize{\textcolor{red}{+4.4}}          & 39.5\footnotesize{\textcolor{red}{+8.1}}          \\
                                      & \checkmark    & \checkmark          & \checkmark        & \textbf{46.3}\footnotesize{\textcolor{red}{+20.9}} & \textbf{52.5}\footnotesize{\textcolor{red}{+21.8}} & \textbf{57.9}\footnotesize{\textcolor{red}{+21.5}} & \textbf{44.7}\footnotesize{\textcolor{red}{+15.9}} & \textbf{37.4}\footnotesize{\textcolor{red}{+13.3}} & \textbf{21.7}\footnotesize{\textcolor{red}{+9.3}} & \textbf{48.7}\footnotesize{\textcolor{red}{+17.3}} \\ \bottomrule
    \end{tabular}%
    }
    \end{table}

\begin{figure}[ht]
	\begin{minipage}[t]{0.48\linewidth}
		\centering
		\includegraphics[width=1.0\textwidth]{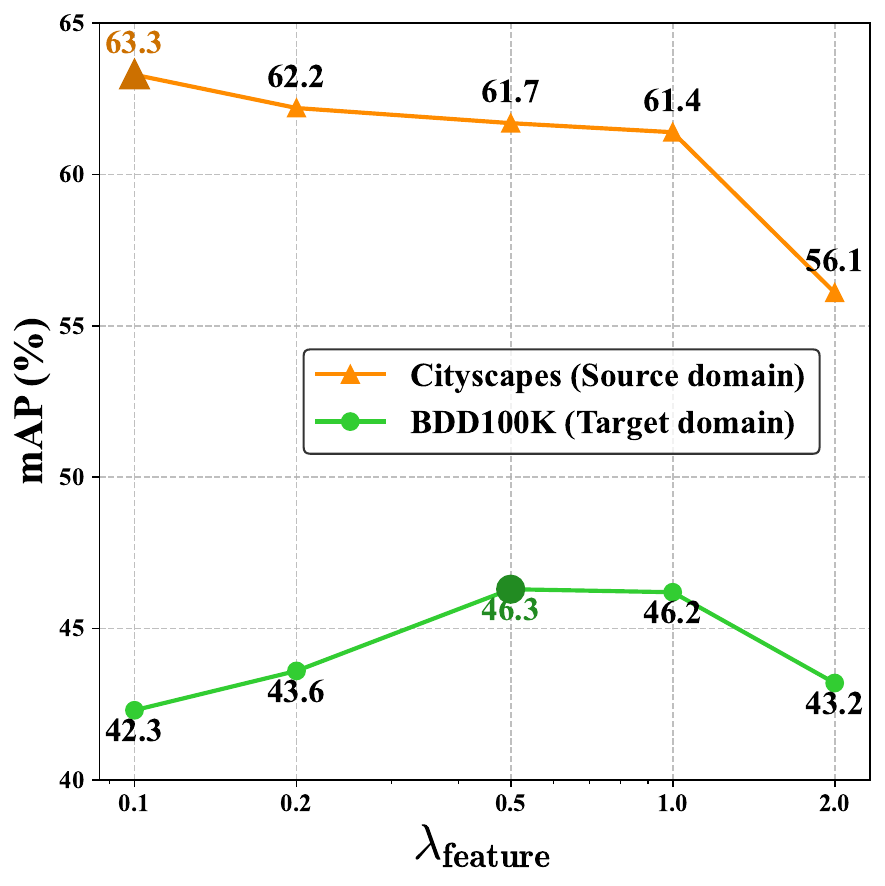}
	\end{minipage}
	\begin{minipage}[t]{0.48\linewidth}
		\centering
		\includegraphics[width=1.0\textwidth]{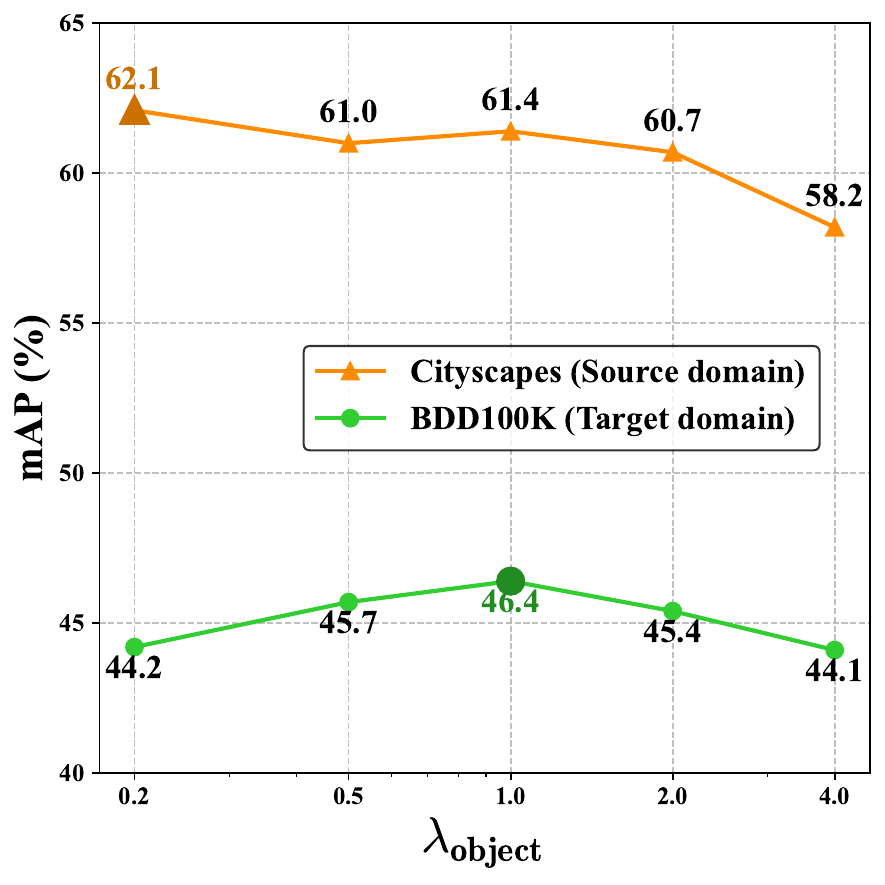}
	\end{minipage}
	\caption{Testing results of $\lambda_{\text{feature}}$ and $\lambda_{\text{object}}$ on source domain and target domain.}
	\label{fig:loss weights}
    \vspace{-8pt}
\end{figure}

\noindent\textbf{Analysis of $\lambda_{\text{feature}}$ and $\lambda_{\text{object}}$} : As shown in Fig.~\ref{fig:loss weights}, we investigate the impact of different weighting factors $\lambda_{\text{feature}}$ and $\lambda_{\text{object}}$. We observe that larger weights lead to improved performance on the target domain while slightly degrading the source domain performance. This trade-off suggests an inherent conflict between the model's ability to fit training data and generalize to unseen target domain data.

\subsection{Model Calibration Performance for DG}
Our diffusion detector demonstrates superior confidence calibration compared to FR-R101 baseline (Fig.~\ref{fig:dece}, Tab.~\ref{tab:dece}). The Diff. Detector reduces D-ECE~\cite{dece} to 8.5\%, 8.7\%, and 5.5\% on BDD100K, Cityscapes, and Clipart datasets respectively. As shown in Fig.~\ref{fig:dece}, our calibration curves more closely follow the diagonal line, indicating better alignment between predicted confidence scores and empirical accuracies across domains.

\subsection{Limitations}
Despite achieving strong results across various benchmarks, our diffusion detector faces efficiency challenges. The large parameter count and multi-step denoising process incur substantial computational costs (Tab.~\ref{tab:diffusion_steps}), limiting its applicability to larger-scale scenarios. 

Moreover, while leveraging diffusion models' generalization capability, performance gains remain limited under severe domain shifts (Tab.~\ref{tab:artistic}). Future work could explore more efficient architectures and effective learning strategies to better handle extreme domain gaps.

\section{Conclusion}
\label{sec:conclusion}

This paper addresses DG detection through two key contributions. First, we propose a diffusion detector that extracts domain-invariant representations by fusing multi-step features during diffusion. Second, to enable other detectors to benefit from such generalization capability, we develop a diffusion-guided detector framework that transfers knowledge to conventional detectors through feature and object level alignment. Extensive evaluations on six domain generalization benchmarks demonstrate substantial improvements across different domains and corruption types. Our work not only provides an effective solution for domain-generalized detection but also opens up new possibilities for leveraging diffusion models to enhance visual recognition robustness.

\clearpage
{
    \small
    \bibliographystyle{ieeenat_fullname}
    \bibliography{main}
}


\clearpage
\setcounter{page}{1}
\appendix

\section{Overview of Supplementary Material}
This supplementary material provides additional experimental results, implementation details, and analysis to support our main paper. The contents are organized as follows:

\begin{itemize}
    \item \textbf{Additional Description for Methods}: Detailed motivations and derivations for our proposed approach, including diffusion features for DG detection, two-level guidance framework, feature alignment, and object-level alignment mechanism (in Sec. \ref{sec: methods_describe}).
    
    \item \textbf{Class-wise Results Analysis}: Detailed per-category performance analysis on Real to Artistic and Diverse Weather benchmarks (in Sec. \ref{sec: Classwise Results}).
    
    \item \textbf{Additional Results of Different SD Versions}: Comprehensive comparison and analysis of different Stable Diffusion versions (SD-1.5, SD-2.1, SD-3-M) across various benchmarks (in Sec. \ref{sec: Stable Diffusion Versions}).

    \item \textbf{Additional Analysis for Results}: In-depth analysis of domain distribution differences and confusion matrix patterns to validate our method's effectiveness (in Sec. \ref{sec: analysis}).
    
    \item \textbf{Visualization of Detection Results}: Qualitative results demonstrating our method's superior performance across different domain generalization scenarios (in Sec. \ref{sec: Visualization}).
\end{itemize}

\noindent\textbf{List of Tables:}
\begin{itemize}
    \item {\color{red}\textbf{Class-wise Results Tables}}:
    \begin{itemize}
        \item Results on Comic dataset (Tab.~\ref{tab:comic})
        \item Results on Watercolor dataset (Tab.~\ref{tab:watercolor})
        \item Results on Diverse Weather benchmark (Tab.~\ref{tab:dwd-classes})
        \item Results on Clipart dataset (Tab.~\ref{tab:clipart})
    \end{itemize}

    \item {\color{red}\textbf{Model Comparison Tables}}:
        \begin{itemize}
            \item Comparison of different SD versions (Tab.~\ref{tab:version})
        \end{itemize}

\end{itemize}

\noindent\textbf{List of Figures:}
\begin{itemize}
    \item {\color{red}\textbf{Analysis Visualizations}}:
        \begin{itemize}
            \item Domain distribution visualization (Fig.~\ref{fig:umap})
            \item Confusion matrices on FoggyCityscapes (Fig.~\ref{fig:confusion matrix_foggy})
            \item Confusion matrices on Clipart (Fig.~\ref{fig:confusion matrix_clipart})
        \end{itemize}
    \item {\color{red}\textbf{Detection Results}}:
        \begin{itemize}
            \item Qualitative results on BDD100K (Fig.~\ref{fig:vis_bdd100k})
            \item Qualitative results on FoggyCityscapes (Fig.~\ref{fig:vis_foggycityscapes})
            \item Qualitative results on Cityscapes (Car) (Fig.~\ref{fig:vis_cityscapes(car)})
            \item Qualitative results on Clipart (Fig.~\ref{fig:vis_clipart})
            \item Qualitative results on Diverse Weather Benchmark (Fig.~\ref{fig:vis_dwd})
            \item Qualitative results on Corruption Benchmark (Fig.~\ref{fig:vis_cityscapes-c})
        \end{itemize}
\end{itemize}

\section{Additional Description for Methods}
\label{sec: methods_describe}
\subsection{Motivations}

\noindent\textbf{Diffusion features for DG detection.}
Domain generalization for object detection requires learning domain-invariant representations without accessing target domain data, which remains challenging due to complex real-world variations. While existing detectors struggle under domain shifts~\cite{dg}, diffusion models have demonstrated unique advantages in handling diverse variations through their progressive denoising process. These models naturally distinguish intrinsic semantic structures from domain-specific variations~\cite{hyperfeatures,tang2023emergent}, building robustness against various perturbations. We leverage these properties for domain-generalized detection: the denoising mechanism filters out domain-specific variations while preserving essential object characteristics~\cite{ddt}, and the multi-scale features provide robust semantic representations that generalize across domains.

\noindent\textbf{Two-level guidance from frozen diffusion detector.}
While directly using diffusion features provides strong generalization capability, it incurs substantial computational overhead. This motivates us to transfer the generalization ability from a frozen diffusion detector to lightweight detectors. We propose a two-level guidance framework to capture both semantic understanding and detection knowledge. At the feature level, we align global feature distributions between diffusion and conventional detectors to learn domain-invariant representations. At the object level, we facilitate task-specific knowledge transfer through shared detection heads following~\cite{crosskd}, enabling precise localization and classification learning from the diffusion teacher.

\noindent\textbf{Feature alignment for heterogeneous detectors.}
Direct feature alignment with MSE leads to suboptimal results due to different magnitude distributions and feature dominance issues~\cite{pkd}. We leverage Pearson Correlation Coefficient (PCC) for feature alignment, which captures relational patterns while being invariant to magnitude differences. By normalizing features before alignment, PCC effectively handles discrepancies between diffusion and conventional detectors, enabling stable knowledge transfer between heterogeneous detector pairs.

\noindent\textbf{Object-level alignment for task-specific knowledge transfer.}
While feature-level alignment helps learn domain-invariant representations, detection-specific knowledge transfer remains challenging due to architectural differences. We propose an object-level alignment scheme that shares detection heads between student and teacher~\cite{crosskd}, providing task-oriented supervision through classification and regression branches. This complementary guidance enables effective knowledge transfer from the diffusion teacher to conventional detectors.

\section{Classwise Results}
\label{sec: Classwise Results}
\noindent\textbf{Results on Real to Artistic:} 
As shown in Tab.~\ref{tab:clipart},~\ref{tab:comic}, and ~\ref{tab:watercolor}, our diffusion detector demonstrates remarkable generalization capability on artistic-style datasets, surpassing both DG and DA methods significantly. On Clipart, our diffusion detector achieves 58.3\% mAP, leading to substantial improvements of 9.0\% and 19.4\% over the previous best DA method AT~\cite{AT} and DG method DivAlign~\cite{Diversification}, respectively. On Comic dataset, our method reaches 51.9\% mAP, exhibiting clear advantages compared to the best DA approach D-ADAPT~\cite{dadapt} at 40.5\% and DG method DivAlign~\cite{Diversification} at 33.2\%. For Watercolor, we achieve 68.4\% mAP, which significantly surpasses the previous best results of 59.9\% from AT~\cite{AT} and 57.4\% from DivAlign~\cite{Diversification}.

However, the diffusion-guided detector shows limited success in bridging extreme domain gaps. On Clipart, Comic, and Watercolor, the diffusion-guided detector (40.8\%, 29.7\%, and 54.2\% respectively) underperforms compared to both recent DG methods (DivAlign: 38.9\%, 33.2\%, and 57.4\%) and DA approaches (AT and D-ADAPT: 49.3\%, 40.5\%, and 59.9\%). While the improvements over baseline remain notable (+13.6\%, +11.6\%, and +12.7\% respectively), the performance gap suggests that transferring the strong generalization capability from diffusion models to conventional detectors remains challenging when facing significant stylistic variations, likely due to the extreme domain shifts in artistic styles that make feature alignment particularly difficult.

\noindent\textbf{Results on Diverse Weather Benchmark:} 
As shown in Tab.~\ref{tab:dwd-classes}, our method demonstrates strong robustness across various weather and lighting conditions. For Daytime-Foggy scenarios, our diffusion guided detector achieves 44.7\% mAP, exceeding the previous best result from UFR~\cite{ufr} by 5.1\%. In Night-Sunny conditions, we obtain 49.1\% mAP, surpassing G-NAS~\cite{gnas} which achieves 45.0\%. The improvement becomes more pronounced in challenging Night-Rainy scenarios, where our diffusion detector reaches 27.8\% mAP, considerably outperforming the previous best of 24.1\% from DivAlign~\cite{Diversification}. Under Dusk-Rainy conditions, we achieve 42.5\% mAP, marking a clear advancement over DivAlign~\cite{Diversification} at 38.1\%. Most notably, the diffusion-guided detector demonstrates consistent improvements over the baseline across all four scenarios, with remarkable margins of +15.9\%, +17.7\%, +9.3\%, and +13.3\%. These comprehensive results not only validate the effectiveness of our knowledge transfer framework in handling natural environmental variations but also confirm our approach's strong capability in enhancing detection generalization across diverse real-world conditions.

\begin{table}[t]
    \centering
    \renewcommand{\arraystretch}{1.1}
    \caption{Real to Artistic DG and DA Results (\%) on Comic (Classwise).}
    \label{tab:comic}
    \setlength{\tabcolsep}{3pt}
    \resizebox{\columnwidth}{!}{%
    \begin{tabular}{l|cccccc>{\columncolor{gray!20}}c}
    \toprule
    \textbf{Methods} & \textbf{\footnotesize{Bike}} & \textbf{\footnotesize{Bird}} & \textbf{\footnotesize{Car}} & \textbf{\footnotesize{Cat}} & \textbf{\footnotesize{Dog}} & \textbf{\footnotesize{Person}} & \textbf{mAP} \\
    \midrule
    \multicolumn{8}{c}{\textit{\textbf{DG methods} (without target data)}} \\
    Div.~\cite{Diversification} \textit{\scriptsize{(\textcolor{darkgreen}{CVPR'24})}} & 41.7 & 12.3 & 29.0 & 13.2 & 20.6 & 36.5 & 25.5 \\
    DivAlign~\cite{Diversification} \textit{\scriptsize{(\textcolor{darkgreen}{CVPR'24})}} & 54.1 & 16.9 & 30.1 & 25.0 & 27.4 & 45.9 & 33.2 \\
    \midrule
    \multicolumn{8}{c}{\textit{\textbf{DA methods} (with unlabeled target data)}} \\
    DA-Faster~\cite{chen2018da-faster} \textit{\scriptsize{(\textcolor{darkgreen}{CVPR'18})}} & 31.1 & 10.3 & 15.5 & 12.4 & 19.3 & 39.0 & 21.2 \\
    SWDA~\cite{saito2019swda} \textit{\scriptsize{(\textcolor{darkgreen}{CVPR'19})}}& 36.4 & 21.8 & 29.8 & 15.1 & 23.5 & 49.6 & 29.4 \\
    STABR~\cite{stabr} \textit{\scriptsize{(\textcolor{darkgreen}{ICCV'19})}}& 50.6 & 13.6 & 31.0 & 7.5 & 16.4 & 41.4 & 26.8 \\
    MCRA~\cite{mcra} \textit{\scriptsize{(\textcolor{darkgreen}{ECCV'20})}}& 47.9 & 20.5 & 37.4 & 20.6 & 24.5 & 50.2 & 33.5 \\
    I3Net~\cite{i3net} \textit{\scriptsize{(\textcolor{darkgreen}{CVPR'21})}}& 47.5 & 19.9 & 33.2 & 11.4 & 19.4 & 49.1 & 30.1 \\
    DBGL~\cite{dbgl} \textit{\scriptsize{(\textcolor{darkgreen}{ICCV'21})}}& 35.6 & 20.3 & 33.9 & 16.4 & 26.6 & 45.3 & 29.7 \\
    D-ADAPT~\cite{dadapt} \textit{\scriptsize{(\textcolor{darkgreen}{ICLR'22})}}& 52.4 & 25.4 & 42.3 & \textbf{43.7} & 25.7 & 53.5 & 40.5 \\
    \midrule
    \multicolumn{8}{c}{\textit{\textbf{Ours (DG settings)}}} \\
    \rowcolor{lightgreen} \textbf{Diff. Detector} \footnotesize{(SD-1.5)} & \textbf{63.3} & \textbf{41.7} & \textbf{58.2} & \underline{31.8} & \textbf{40.9} & \textbf{75.3} & \textbf{51.9} \\
    \textbf{Diff. Detector} \footnotesize{(SD-2.1)} & \underline{61.1} & \underline{35.7} & \underline{53.6} & 23.2 & \underline{35.0} & \underline{71.2} & \underline{46.6} \\ 
    \textbf{Diff. Guided} \footnotesize{(SD-1.5)} & 47.6 & 21.0 & 35.3 & 9.1 & 21.6 & 43.5 & 29.7\footnotesize{\textcolor{red}{+11.6}} \\
    \textbf{Diff. Guided} \footnotesize{(SD-2.1)} & 46.4 & 13.2 & 24.2 & 7.5 & 12.3 & 35.8 & 24.9\footnotesize{\textcolor{red}{+6.8}} \\
    \bottomrule
    \end{tabular}%
    }
\end{table}


\begin{table}[t]
    \centering
    \renewcommand{\arraystretch}{1.0}
    \caption{Real to Artistic DG and DA Results (\%) on Watercolor (Classwise).}
    \label{tab:watercolor}
    \setlength{\tabcolsep}{3pt}
    \resizebox{\columnwidth}{!}{%
    \begin{tabular}{l|cccccc>{\columncolor{gray!20}}c}
    \toprule
    \textbf{Methods} & \textbf{\footnotesize{Bike}} & \textbf{\footnotesize{Bird}} & \textbf{\footnotesize{Car}} & \textbf{\footnotesize{Cat}} & \textbf{\footnotesize{Dog}} & \textbf{\footnotesize{Person}} & \textbf{mAP} \\
    \midrule
    \multicolumn{8}{c}{\textit{\textbf{DG methods} (without target data)}} \\
    Div.~\cite{Diversification} \textit{\scriptsize{(\textcolor{darkgreen}{CVPR'24})}}& 87.1 & 51.7 & 53.6 & 35.1 & 23.6 & 63.6 & 52.5 \\
    DivAlign~\cite{Diversification} \textit{\scriptsize{(\textcolor{darkgreen}{CVPR'24})}}& 90.4 & 51.8 & 51.9 & 43.9 & 35.9 & 70.2 & 57.4 \\
    \midrule
    \multicolumn{8}{c}{\textit{\textbf{DA methods} (with unlabeled target data)}} \\
    SWDA~\cite{saito2019swda} \textit{\scriptsize{(\textcolor{darkgreen}{CVPR'19})}} & 82.3 & 55.9 & 46.5 & 32.7 & 35.5 & 66.7 & 53.3 \\
    MCRA~\cite{mcra} \textit{\scriptsize{(\textcolor{darkgreen}{ECCV'20})}} & 87.9 & 52.1 & 51.8 & 41.6 & 33.8 & 68.8 & 56.0 \\
    UMT~\cite{umt} \textit{\scriptsize{(\textcolor{darkgreen}{CVPR'21})}} & 88.2 & 55.3 & 51.7 & 39.8 & 43.6 & 69.9 & 58.1 \\
    IIOD~\cite{iiod} \textit{\scriptsize{(\textcolor{darkgreen}{TPAMI'21})}}& 95.8 & 54.3 & 48.3 & 42.4 & 35.1 & 65.8 & 56.9 \\
    I3Net~\cite{i3net} \textit{\scriptsize{(\textcolor{darkgreen}{CVPR'21})}} & 81.1 & 49.3 & 46.2 & 35.0 & 31.9 & 65.7 & 51.5 \\
    SADA~\cite{sada} \textit{\scriptsize{(\textcolor{darkgreen}{IJCV'21})}}& 82.9 & 54.6 & 52.3 & 40.5 & 37.7 & 68.2 & 56.0 \\
    CDG~\cite{cdg} \textit{\scriptsize{(\textcolor{darkgreen}{CVPR'19})}}& 97.7 & 53.1 & 52.1 & 47.3 & 38.7 & 68.9 & 59.7 \\
    VDD~\cite{saito2019swda} \textit{\scriptsize{(\textcolor{darkgreen}{AAAI'21})}}& 90.0 & 56.6 & 49.2 & 39.5 & 38.8 & 65.3 & 56.6 \\
    DBGL~\cite{dbgl} \textit{\scriptsize{(\textcolor{darkgreen}{ICCV'21})}} & 83.1 & 49.3 & 50.6 & 39.8 & 38.7 & 61.3 & 53.8 \\
    AT~\cite{AT} \textit{\scriptsize{(\textcolor{darkgreen}{CVPR'22})}}& 93.6 & 56.1 & 58.9 & 37.3 & 39.6 & \underline{73.8} & 59.9 \\
    LODS~\cite{lods} \textit{\scriptsize{(\textcolor{darkgreen}{CVPR'22})}}& 95.2 & 53.1 & 46.9 & 37.2 & \underline{47.6} & 69.3 & 58.2 \\
    \midrule
    \multicolumn{8}{c}{\textit{\textbf{Ours (DG settings)}}} \\
    \rowcolor{lightgreen} \textbf{Diff. Detector} \footnotesize{(SD-1.5)} & \textbf{99.8} & \textbf{70.3} & \textbf{57.5} & \textbf{49.8} & \textbf{51.0} & \textbf{82.0} & \textbf{68.4} \\
    \textbf{Diff. Detector} \footnotesize{(SD-2.1)} & 91.1 & \underline{65.9} & \underline{55.7} & \underline{47.6} & 39.1 & 73.4 & \underline{62.1} \\ 
    \textbf{Diff. Guided} \footnotesize{(SD-1.5)} & 90.1 & 51.0 & 48.5 & 40.2 & 28.9 & 66.7 & 54.2\footnotesize{\textcolor{red}{+12.7}} \\
    \textbf{Diff. Guided} \footnotesize{(SD-2.1)} & \underline{99.6} & 48.4 & 49.1 & 28.4 & 23.4 & 54.2 & 50.6\footnotesize{\textcolor{red}{+9.1}} \\
    \bottomrule
    \end{tabular}%
    }
\end{table}

\begin{table*}[ht]
    \centering
    \renewcommand{\arraystretch}{1.0}
    \caption{Generalization detection Results (\%) on Diverse Weather benchmark (Classwise).}
    \label{tab:dwd-classes}
    \setlength{\tabcolsep}{6pt}
    \resizebox{\textwidth}{!}{%
    \begin{tabular}{l|ccccccc>{\columncolor{gray!20}}c|ccccccc>{\columncolor{gray!20}}c}
    \toprule
                                  & \multicolumn{8}{c|}{\textbf{Daytime-Foggy}}                                  & \multicolumn{8}{c}{\textbf{Night-Sunny}}                                      \\
                                  \textbf{Methods} & \textbf{Bus} & \textbf{Bike} & \textbf{Car} & \textbf{Motor} & \textbf{Person} & \textbf{Rider} & \textbf{Truck} & \textbf{mAP} & \textbf{Bus} & \textbf{Bike} & \textbf{Car} & \textbf{Motor} & \textbf{Person} & \textbf{Rider} & \textbf{Truck} & \textbf{mAP} \\ \midrule
    IBN-Net \cite{ibnnet} \textit{\scriptsize{(\textcolor{darkgreen}{CVPR'18})}}                      & 29.9 & 26.1 & 44.5 & 24.4  & 26.2   & 33.5  & 22.4  & 29.6 & 37.8 & 27.3 & 49.6 & 15.1  & 29.2   & 27.1  & 38.9  & 32.1 \\
    SW \cite{sw} \textit{\scriptsize{(\textcolor{darkgreen}{ICCV'19})}}                          & 30.6 & 26.2 & 44.6 & 25.1  & 30.7   & 34.6  & 23.6  & 30.8 & 38.7 & 29.2 & 49.8 & 16.6  & 31.5   & 28.0  & 40.2  & 33.4 \\
    IterNorm \cite{iternorm} \textit{\scriptsize{(\textcolor{darkgreen}{CVPR'19})}}                    & 29.7 & 21.8 & 42.4 & 24.4  & 26.0   & 33.3  & 21.6  & 28.5 & 38.5 & 23.5 & 38.9 & 15.8  & 26.6   & 25.9  & 38.1  & 29.6 \\
    ISW \cite{isw} \textit{\scriptsize{(\textcolor{darkgreen}{CVPR'21})}}                         & 29.5 & 26.4 & 49.2 & 27.9  & 30.7   & 34.8  & 24.0  & 31.8 & 38.5 & 28.5 & 49.6 & 15.4  & 31.9   & 27.5  & 41.3  & 33.2 \\
    CDSD \cite{cdsd} \textit{\scriptsize{(\textcolor{darkgreen}{CVPR'22})}}                        & 32.9 & 28.0 & 48.8 & 29.8  & 32.5   & 38.2  & 24.1  & 33.5 & 40.6 & 35.1 & 50.7 & 19.7  & 34.7   & 32.1  & 43.4  & 36.6 \\
    CLIPGap \cite{clip_gap} \textit{\scriptsize{(\textcolor{darkgreen}{CVPR'23})}}                       & 36.1 & 34.3 & 58.0 & 33.1  & 39.0   & 43.9  & 25.1  & 38.5 & 37.7 & 34.3 & 58.0 & 19.2  & 37.6   & 28.5  & 42.9  & 36.9 \\
    SRCD \cite{srcd} \textit{\scriptsize{(\textcolor{darkgreen}{TNNLS'24})}}                        & 36.4 & 30.1 & 52.4 & 31.3  & 33.4   & 40.1  & 27.7  & 35.9 & 43.1 & 32.5 & 52.3 & 20.1  & 34.8   & 31.5  & 42.9  & 36.7 \\
    G-NAS \cite{gnas} \textit{\scriptsize{(\textcolor{darkgreen}{AAAI'24})}}                       & 32.4 & 31.2 & 57.7 & 31.9  & 38.6   & 38.5  & 24.5  & 36.4 & 46.9 & 40.5 & 67.5 & 26.5  & 50.7   & 35.4  & 47.8  & 45.0 \\
    OA-DG \cite{oamix} \textit{\scriptsize{(\textcolor{darkgreen}{AAAI'24})}}                       & -    & -    & -    & -     & -      & -     & -     & 38.3 & -    & -    & -    & -     & -      & -     & -     & 38.0 \\
    DivAlign \cite{Diversification} \textit{\scriptsize{(\textcolor{darkgreen}{CVPR'24})}}                    & -    & -    & -    & -     & -      & -     & -     & 37.2 & -    & -    & -    & -     & -      & -     & -     & 42.5 \\
    UFR \cite{ufr} \textit{\scriptsize{(\textcolor{darkgreen}{CVPR'24})}}                         & 36.9 & 35.8 & 61.7 & 33.7  & 39.5   & 42.2  & 27.5  & 39.6 & 43.6 & 38.1 & 66.1 & 14.7  & 49.1   & 26.4  & 47.5  & 40.8 \\ \midrule \midrule
    \textbf{Diff. Detector} \footnotesize{(SD-1.5)}      & 37.5 & 32.4 & 67.9 & 35.6  & 48.3   & 44.6  & \textbf{37.1}  & 43.3 & \underline{49.6} & 42.1 & 70.5 & 21.4  & 54.5   & 38.2  & \underline{52.6}  & 47.0 \\
    \textbf{Diff. Detector} \footnotesize{(SD-2.1)}      & 36.4 & \textbf{36.7} &68.8 & 36.6  & \textbf{51.5}   & 49.1  & 32.9  & 44.6 & 48.2 & 39.6 & 69.2 & 22.8  & 55.4   & 37.7  & 51.6  & 46.4 \\ \midrule
    \textbf{Diff. Guided} \footnotesize{(SD-1.5)} & \textbf{39.3} & 35.8 & \textbf{69.4} & \textbf{37.7}  & \underline{48.8}   & \textbf{49.7}  & 32.3  & \underline{44.7}\footnotesize{\textcolor{red}{+15.9}} & \underline{51.0} & \underline{42.8} & \underline{72.2} & \underline{27.5}  & \underline{55.9}   & \underline{39.5}  & 52.0  & \underline{48.6}\footnotesize{\textcolor{red}{+17.2}} \\
    \rowcolor{lightgreen} \textbf{Diff. Guided} \footnotesize{(SD-2.1)} & \underline{38.8} & \underline{36.4} & \underline{68.9} & \underline{37.4}  & 48.6   & 49.6  & \underline{33.4}  & \textbf{44.7}\footnotesize{\textcolor{red}{+15.9}} & \textbf{51.3} & \textbf{43.6} & \textbf{72.3} & \textbf{27.6}  & \textbf{56.2}   & \textbf{40.2}  & \textbf{53.7}  & \textbf{49.1}\footnotesize{\textcolor{red}{+17.7}} \\ \midrule
    & \multicolumn{8}{c|}{\textbf{Night-Rainy}}                                  & \multicolumn{8}{c}{\textbf{Dusk-Rainy}}                                      \\ \midrule
    IBN-Net \cite{ibnnet} \textit{\scriptsize{(\textcolor{darkgreen}{CVPR'18})}}                      & 24.6 & 10.0 & 28.4 & 0.9  & 8.3    & 9.8   & 18.1  & 14.3 & 37.0 & 14.8 & 50.3 & 11.4  & 17.3   & 13.3  & 38.4  & 26.1 \\
    SW \cite{sw} \textit{\scriptsize{(\textcolor{darkgreen}{ICCV'19})}}                           & 22.3 & 7.8  & 27.6 & 0.2  & 10.3   & 10.0  & 17.7  & 13.7 & 35.2 & 16.7 & 50.1 & 10.4  & 20.1   & 13.0  & 38.8  & 26.3 \\
    IterNorm \cite{iternorm} \textit{\scriptsize{(\textcolor{darkgreen}{CVPR'19})}}                     & 21.4 & 6.7  & 22.0 & 0.9  & 9.1    & 10.6  & 17.6  & 12.6 & 32.9 & 14.1 & 38.9 & 11.0  & 15.5   & 11.6  & 35.7  & 22.8 \\
    ISW \cite{isw} \textit{\scriptsize{(\textcolor{darkgreen}{CVPR'21})}}                          & 22.5 & 11.4 & 26.9 & 0.4  & 9.9    & 9.8   & 17.5  & 14.1 & 34.7 & 16.0 & 50.0 & 11.1  & 17.8   & 12.6  & 38.8  & 25.9 \\
    CDSD \cite{cdsd} \textit{\scriptsize{(\textcolor{darkgreen}{CVPR'22})}}                         & 24.4 & 11.6 & 29.5 & 0.4  & 10.5   & 11.4  & 19.2  & 15.3 & 37.1 & 19.6 & 50.9 & 13.4  & 19.7   & 16.3  & 40.7  & 28.2 \\
    CLIPGap \cite{clip_gap} \textit{\scriptsize{(\textcolor{darkgreen}{CVPR'23})}}                      & 28.6 & 12.1 & 36.1 & 9.2  & 12.3   & 9.6   & 22.9  & 18.7 & 37.8 & 22.8 & 60.7 & 16.8  & 26.8   & 18.7  & 42.4  & 32.3 \\
    SRCD \cite{srcd} \textit{\scriptsize{(\textcolor{darkgreen}{TNNLS'24})}}                         & 26.5 & \underline{12.9} & 32.4 & 0.8  & 10.2   & \underline{12.5}  & 24.0  & 17.0 & 39.5 & 21.4 & 50.6 & 11.9  & 20.1   & 17.6  & 40.5  & 28.8 \\
    G-NAS \cite{gnas} \textit{\scriptsize{(\textcolor{darkgreen}{AAAI'24})}}                        & 28.6 & 9.8  & 38.4 & 0.1  & 13.8   & 9.8   & 21.4  & 17.4 & \underline{44.6} & 22.3 & 66.4 & 14.7  & 32.1   & 19.6  & 45.8  & 35.1 \\
    OA-DG \cite{oamix} \textit{\scriptsize{(\textcolor{darkgreen}{AAAI'24})}}                        & -    & -    & -    & -    & -      & -     & -     & 16.8 & -    & -    & -    & -     & -      & -     & -     & 33.9 \\
    DivAlign \cite{Diversification} \textit{\scriptsize{(\textcolor{darkgreen}{CVPR'24})}}                     & -    & -    & -    & -    & -      & -     & -     & \underline{24.1} & -    & -    & -    & -     & -      & -     & -     & 38.1 \\
    UFR \cite{ufr} \textit{\scriptsize{(\textcolor{darkgreen}{CVPR'24})}}                          & 29.9 & 11.8 & 36.1 & \underline{9.4}  & 13.1   & 10.5  & 23.3  & 19.2 & 37.1 & 21.8 & 67.9 & 16.4  & 27.4   & 17.9  & 43.9  & 33.2 \\ \midrule \midrule
    \rowcolor{lightgreen} \textbf{Diff. Detector} \footnotesize{(SD-1.5)}      & \textbf{42.0} & \textbf{15.0} & \textbf{53.6} & 6.5  & \textbf{26.2}   & \textbf{13.8}  & \textbf{37.5}  & \textbf{27.8} & \textbf{49.7} & \underline{27.9} & \textbf{74.9} & \underline{18.2}  & \textbf{45.5}   & \textbf{24.5}  & \textbf{56.8}  & \textbf{42.5} \\
    \textbf{Diff. Detector} \footnotesize{(SD-2.1)}      & 30.1 & 11.3 & 46.1 & \textbf{10.2}  & \underline{24.1}   & 9.2   & \underline{31.5}  & 23.2 & \underline{44.6} & \textbf{30.6} & 73.5 & \textbf{22.1}  & \underline{44.4}   & 20.1  & \underline{55.6}  & \underline{41.6} \\ \midrule
    \textbf{Diff. Guided} \footnotesize{(SD-1.5)} & \underline{35.4} & 12.7 & \underline{46.2} & 3.2  & 13.8   & 10.7  & 29.7  & 21.7\footnotesize{\textcolor{red}{+9.3}} & 43.1 & 23.9 & \underline{73.6} & 13.4  & 33.2   & \underline{22.1}  & 52.3  & 37.4\footnotesize{\textcolor{red}{+13.3}} \\
    \textbf{Diff. Guided} \footnotesize{(SD-2.1)} & 34.4 & 7.8  & 43.3 & 2.2  & 14.3   & 7.5   & 30.3  & 20.8\footnotesize{\textcolor{red}{+8.4}} & \underline{44.6} & 22.5 & 73.1 & 15.7  & 31.7   & 19.3  & 52.6  & 37.3\footnotesize{\textcolor{red}{+13.2}} \\ \bottomrule
    \end{tabular}%
    }
\end{table*}

\begin{table*}[ht]
    \centering
    \renewcommand{\arraystretch}{1.25}
    \caption{Real to Artistic DG and DA Results (\%) on Clipart (Classwise).}
    \setlength{\tabcolsep}{2pt}
    \label{tab:clipart}
    \resizebox{\textwidth}{!}{%
    \begin{tabular}{l|cccccccccccccccccccc>{\columncolor{gray!20}}c}
    \toprule
    \textbf{Methods} & \textbf{aero.} & \textbf{bike} & \textbf{bird} & \textbf{boat} & \textbf{bottle} & \textbf{bus} & \textbf{car} & \textbf{cat} & \textbf{chair} & \textbf{cow} & \textbf{table} & \textbf{dog} & \textbf{horse} & \textbf{bike} & \textbf{psn.} & \textbf{plant.} & \textbf{sheep} & \textbf{sofa} & \textbf{train} & \textbf{tv} & \textbf{mAP} \\
    \midrule
    \multicolumn{22}{c}{\textit{\textbf{DG methods} (without target data)}} \\
    Div.~\cite{Diversification} \textit{\scriptsize{(\textcolor{darkgreen}{CVPR'24})}} & 29.3 & 50.9 & 23.4 & 35.3 & 45.3 & 49.8 & 33.4 & 10.6 & 43.3 & 22.3 & 31.6 & 4.5 & 32.9 & 51.9 & 40.2 & 51.1 & 18.2 & 29.6 & 42.3 & 28.5 & 33.7 \\
    DivAlign~\cite{Diversification} \textit{\scriptsize{(\textcolor{darkgreen}{CVPR'24})}} & 34.4 & 64.4 & 22.7 & 27.0 & 45.6 & 59.2 & 32.9 & 7.0 & 46.8 & 55.8 & 28.9 & 14.5 & 44.4 & 58.0 & 55.2 & 52.1 & 14.8 & 38.4 & 42.5 & 33.9 & 38.9 \\
    \midrule
    \multicolumn{22}{c}{\textit{\textbf{DA methods} (with unlabeled target data)}} \\
    AT~\cite{AT} \textit{\scriptsize{(\textcolor{darkgreen}{CVPR'22})}}& 33.8 & 60.9 & 38.6 & 49.4 & 52.4 & 53.9 & 56.7 & 7.5 & 52.8 & \textbf{63.5} & 34.0 & 25.0 & 62.2 & 72.1 & 77.2 & 57.7 & 27.2 & 52.0 & 55.7 & 54.1 & 49.3 \\
    D-ADAPT~\cite{dadapt} \textit{\scriptsize{(\textcolor{darkgreen}{ICLR'22})}} & 56.4 & 63.2 & 42.3 & 40.9 & 45.3 & \underline{77.0} & 48.7 & \underline{25.4} & 44.3 & 58.4 & 31.4 & 24.5 & 47.1 & 75.3 & 69.3 & 43.5 & \underline{27.9} & 34.1 & \underline{60.7} & \textbf{64.0} & 49.0 \\
    TIA~\cite{tia} \textit{\scriptsize{(\textcolor{darkgreen}{CVPR'22})}} & 42.2 & \underline{66.0} & 36.9 & 37.3 & 43.7 & 71.8 & 49.7 & 18.2 & 44.9 & 58.9 & 18.2 & \underline{29.1} & 40.7 & \underline{87.8} & 67.4 & 49.7 & 27.4 & 27.8 & 57.1 & 50.6 & 46.3 \\
    CIGAR~\cite{CIGAR} \textit{\scriptsize{(\textcolor{darkgreen}{CVPR'23})}} & 35.2 & 55.0 & 39.2 & 30.7 & \textbf{60.1} & 58.1 & 46.9 & \textbf{31.8} & 47.0 & \underline{61.0} & 21.8 & 26.7 & 44.6 & 52.4 & 68.5 & 54.4 & \textbf{31.3} & 38.8 & 56.5 & \underline{63.5} & 46.2 \\
    CMT~\cite{cao2023cmt} \textit{\scriptsize{(\textcolor{darkgreen}{CVPR'23})}} & 39.8 & 56.3 & 38.7 & 39.7 & \underline{60.0} & 35.0 & 56.0 & 7.1 & 60.1 & 60.4 & 35.8 & 28.1 & \textbf{67.8} & 84.5 & 80.1 & 55.5 & 20.3 & 32.8 & 42.3 & 38.2 & 47.0 \\
    \midrule
    \multicolumn{22}{c}{\textit{\textbf{Ours (DG settings)}}} \\
    \rowcolor{lightgreen}\textbf{Diff. Detector} \footnotesize{(SD-1.5)} & \underline{63.7} & \textbf{86.1} & \textbf{49.8} & \underline{56.5} & 52.9 & 50.9 & \textbf{67.3} & 19.7 & \textbf{74.7} & 34.3 & \textbf{57.7} & \textbf{41.9} & \underline{63.2} & \textbf{89.4} & \textbf{89.6} & 59.8 & 23.5 & \textbf{64.9} & \textbf{65.9} & 55.2 & \textbf{58.3} \\
    \textbf{Diff. Detector} \footnotesize{(SD-2.1)} & \textbf{65.5} & 61.7 & \underline{49.5} & \textbf{58.7} & 59.8 & 34.2 & \underline{63.6} & 20.4 & \underline{72.9} & 22.2 & \underline{47.1} & 28.5 & 51.2 & 82.3 & \underline{87.0} & \textbf{61.7} & 20.6 & \underline{57.9} & 44.6 & 44.2 & \underline{51.7} \\ 
    \textbf{Diff. Guided} \footnotesize{(SD-1.5)} & 19.3 & 57.8 & 28.4 & 37.4 & 57.8 & \textbf{81.3} & 46.3 & 3.8 & 57.8 & 27.2 & 28.3 & 19.6 & 42.5 & 50.9 & 57.8 & \underline{59.8} & 15.6 & 36.0 & 37.7 & 50.5 & 40.8\footnotesize{\textcolor{red}{+13.6}} \\
    \textbf{Diff. Guided} \footnotesize{(SD-2.1)} & 25.6 & 40.2 & 26.2 & 25.7 & 44.8 & 72.9 & 34.8 & 3.8 & 46.3 & 14.0 & 26.6 & 7.5 & 27.2 & 57.1 & 48.4 & 56.4 & 6.8 & 25.3 & 24.5 & 39.2 & 32.7\footnotesize{\textcolor{red}{+5.5}} \\
    \bottomrule
    \end{tabular}%
    }
    \end{table*}
\clearpage

\section{Additional Results of Different Stable Diffusion Versions}
\label{sec: Stable Diffusion Versions}

\noindent\textbf{Performance Comparison of Different SD Versions:} 
Experimental results in Tab.~\ref{tab:version} demonstrate varying performance among Stable Diffusion versions across different scenarios. SD-1.5 consistently achieves superior performance, particularly in adverse weather (50.1\% for foggy, 58.2\% for rainy) and artistic style transfer (58.3\%, 51.9\%, 68.4\% for Clipart, Comic, Watercolor). While SD-2.1 maintains competitive performance and achieves 64.5\% accuracy in Cityscapes car detection, it shows performance gaps of 6.6\%, 5.3\%, and 6.3\% compared to SD-1.5 in artistic style transfer. SD-3-M shows significantly lower performance, with substantial degradation in artistic style transfer (28.7\%, 24.1\%, 45.0\%) and diverse weather conditions (10.4\% lower than SD-1.5).

\noindent\textbf{Analysis of Architecture Differences:}
The inferior performance of SD-3-M primarily stems from its architectural differences. Unlike SD-1.5 and SD-2.1 with UNet~\cite{unet} architecture that produces multi-scale hierarchical features, SD-3-M with transformer-based structure~\cite{sd3} outputs fixed-dimensional feature maps. This limitation affects its ability to capture fine-grained spatial information crucial for object detection, particularly impacting performance across diverse domains.

\noindent\textbf{Ongoing Research:}
We are currently conducting extensive experiments to improve the cross-domain detection performance of SD-3-M. Our ongoing research focuses on developing effective methods to leverage the intermediate features of SD-3-M, aiming to fully utilize its strong semantic understanding capabilities while addressing the challenges in dense prediction tasks. The experimental results and detailed analysis will be reported in future work.

\begin{table*}[ht]
    \centering
    \caption{Testing Results of Diffusion Detector with Different Stable Diffusion versions. \textbf{SD-1.5}: Stable Diffusion v1.5, \textbf{SD-2.1}: Stable Diffusion v2.1, \textbf{SD-3-M}: Stable Diffusion v3 Medium. \textbf{Foggy}: FoggyCityscapes, \textbf{Rainy}: RainCityscapes. In Diverse Weather benchmark: \textbf{DF} (Daytime-Foggy), \textbf{DR} (Dusk-Rainy), \textbf{NR} (Night-Rainy), \textbf{NS} (Night-Sunny).}
    \label{tab:version}
    \renewcommand{\arraystretch}{1.2}
    \resizebox{2.0\columnwidth}{!}{%
    \begin{tabular}{c|c|cc|cc|ccc|cccc}
    \toprule
                    & \textbf{Cross Camera} & \multicolumn{2}{c|}{\textbf{Adverse Weather}} & \multicolumn{2}{c|}{\textbf{Synthetic to Real}} & \multicolumn{3}{c|}{\textbf{Real to Artistic}} & \multicolumn{4}{c}{\textbf{Diverse Weather benchmark}} \\ \midrule
    \textbf{Version}         & BDD100K               & Foggy       & Rainy       & Cityscapes (car)         & BDD100K (car)        & Clipart       & Comic       & Watercolor       & DF & DR & NR & NS \\ \midrule
    \textbf{SD-1.5} & \textbf{46.6}        & \textbf{50.1}         & \underline{58.2}        & \underline{62.8}            & \textbf{64.4}        & \textbf{58.3} & \textbf{51.9} & \textbf{68.4}    & \underline{43.3} & \textbf{42.5} & \textbf{27.8} & \textbf{47.0} \\
    \textbf{SD-2.1} & \underline{45.8}     & \underline{48.3}      & 56.1     & \textbf{64.5}              & \underline{64.1}     & \underline{51.7} & \underline{46.6} & \underline{62.1} & \textbf{44.6} & \underline{41.6} & \underline{23.2} & \underline{46.4} \\
    \textbf{SD-3-M} & 40.4                 & 46.1                  & \textbf{59.1}                 & 59.7                       & 54.2                 & 28.7           & 24.1          & 45.0              & 36.0          & 30.5          & 15.9           & 32.8           \\ \bottomrule
    \end{tabular}%
    }
    \end{table*}

\begin{figure*}[hb]
    \centering
    \includegraphics[width=1.0\textwidth]{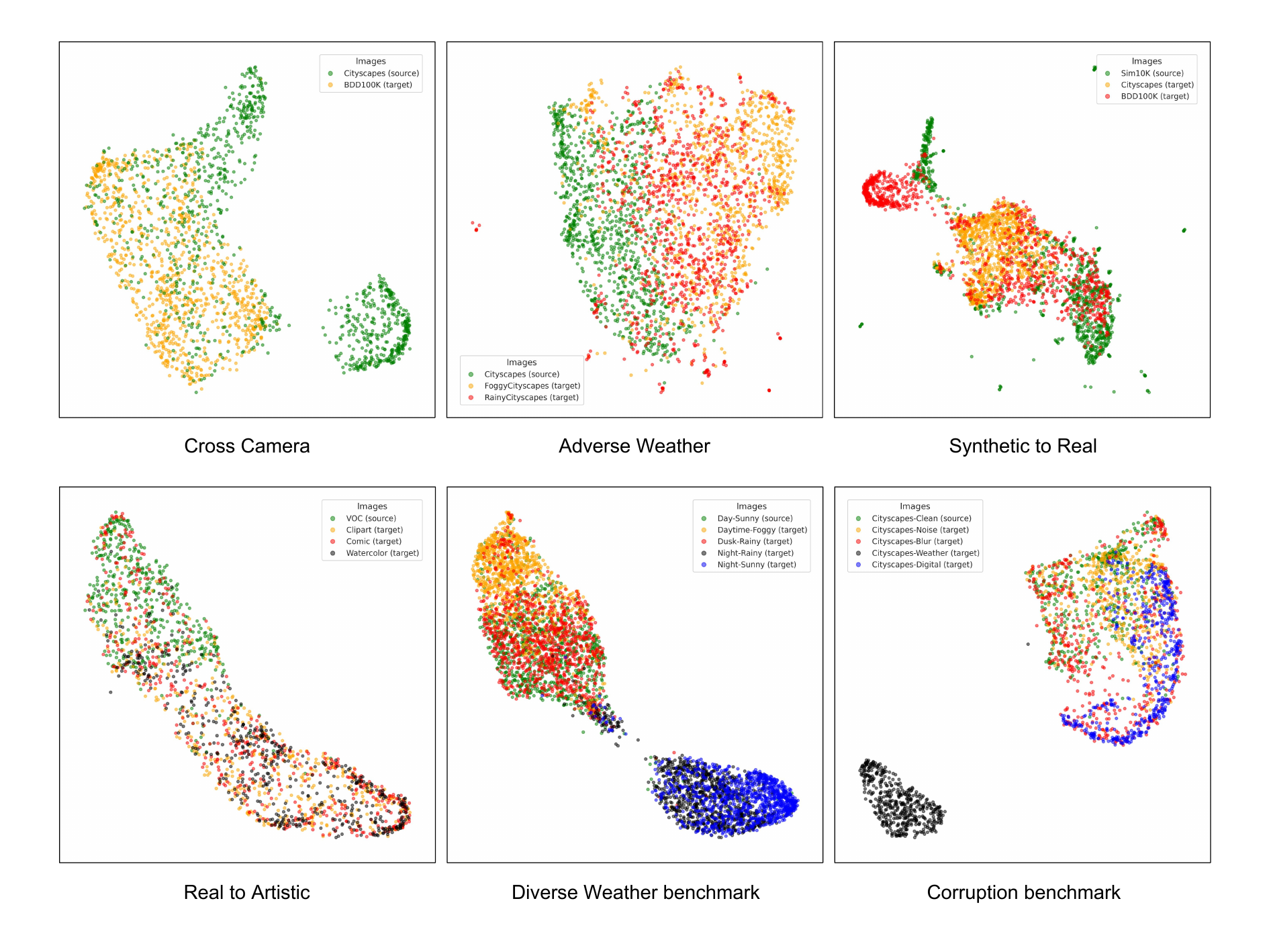}
    \caption{Image-level distribution visualization using UMAP \cite{umap} on six domain generalization benchmarks.}
    \label{fig:umap}
\end{figure*}
\begin{figure*}[h]
    \centering
    \includegraphics[width=1.0\textwidth]{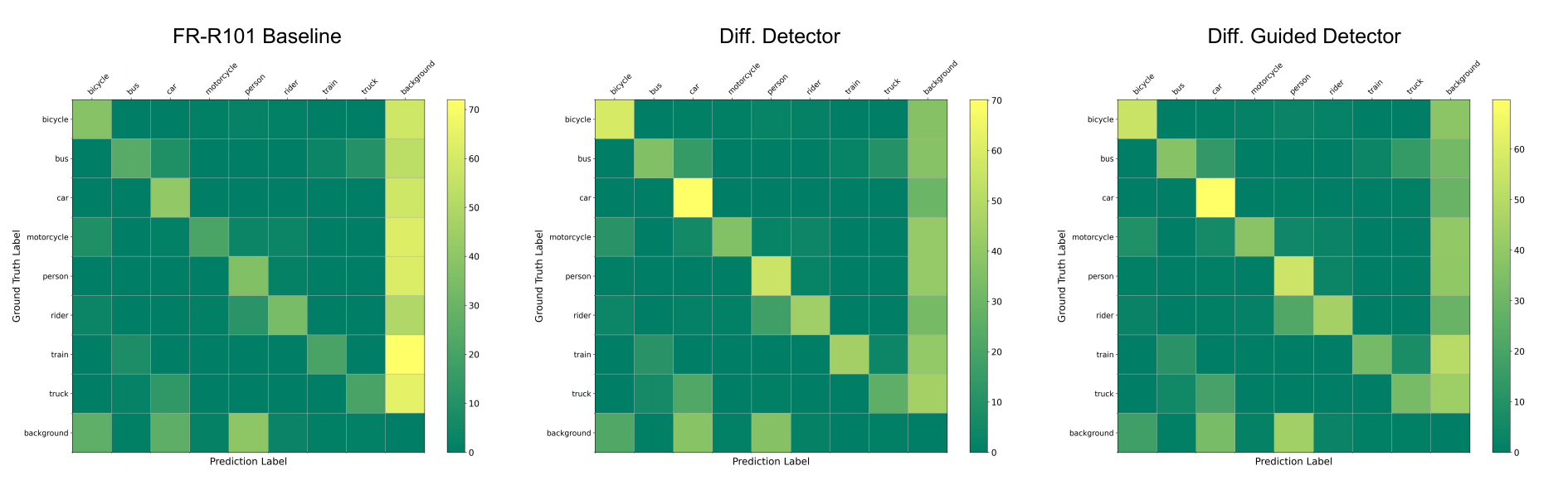}
    \caption{Confusion matrix of \textbf{Baseline} (left), \textbf{Diff. Detector} (middle), and \textbf{Diff. Guided Detector} (right) on FoggyCityscapes.}
    \label{fig:confusion matrix_foggy}
\end{figure*}

\begin{figure*}[h]
    \centering
    \includegraphics[width=1.0\textwidth]{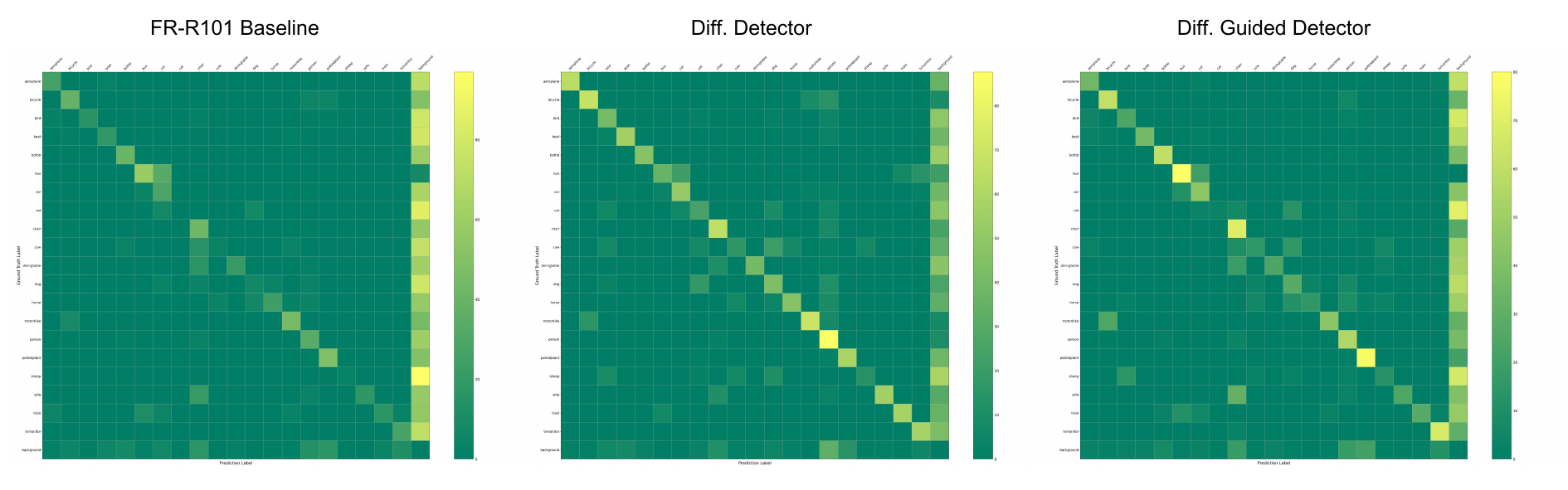}
    \caption{Confusion matrix of \textbf{Baseline} (left), \textbf{Diff. Detector} (middle), and \textbf{Diff. Guided Detector} (right) on Clipart.}
    \label{fig:confusion matrix_clipart}
\end{figure*}
\section{Additional Analysis for Results}
\label{sec: analysis}
\subsection{Visualization of Domain Distribution Differences}
\noindent\textbf{Distribution Analysis:}
As visualized in Fig.~\ref{fig:umap}, significant distribution gaps exist between source and target domains across different benchmarks. The diverse scenarios including cross-camera, adverse weather, synthetic-to-real transfer, artistic style transfer, and various weather conditions all demonstrate distinct distribution separations between source and target domains. These distribution discrepancies explain the challenges faced by conventional detectors when deploying across domains, highlighting the necessity of robust domain-generalized detection approaches.

\subsection{Confusion Matrix Error Analysis}
\noindent\textbf{Analysis of Confusion Matrices:}
As shown in Fig.~\ref{fig:confusion matrix_foggy} and~\ref{fig:confusion matrix_clipart}, the confusion matrices reveal that false negatives (missed detections) are the primary factor affecting detection performance in the baseline detector. Our proposed Diff. Detector significantly reduces the probability of missed detections, as evidenced by the stronger diagonal patterns in both FoggyCityscapes and Clipart scenarios. 

Through our designed feature and object alignment mechanism, the Diff. Guided Detector successfully inherits the robust detection capability from Diff. Detector, showing similar improvements in reducing missed detections. The enhanced diagonal patterns in confusion matrices validate the effectiveness of our knowledge transfer framework in improving cross-domain generalization performance.

\section{Visualization of Detection Results}
\label{sec: Visualization}
\begin{figure*}[h]
    \centering
    \includegraphics[width=1.0\textwidth]{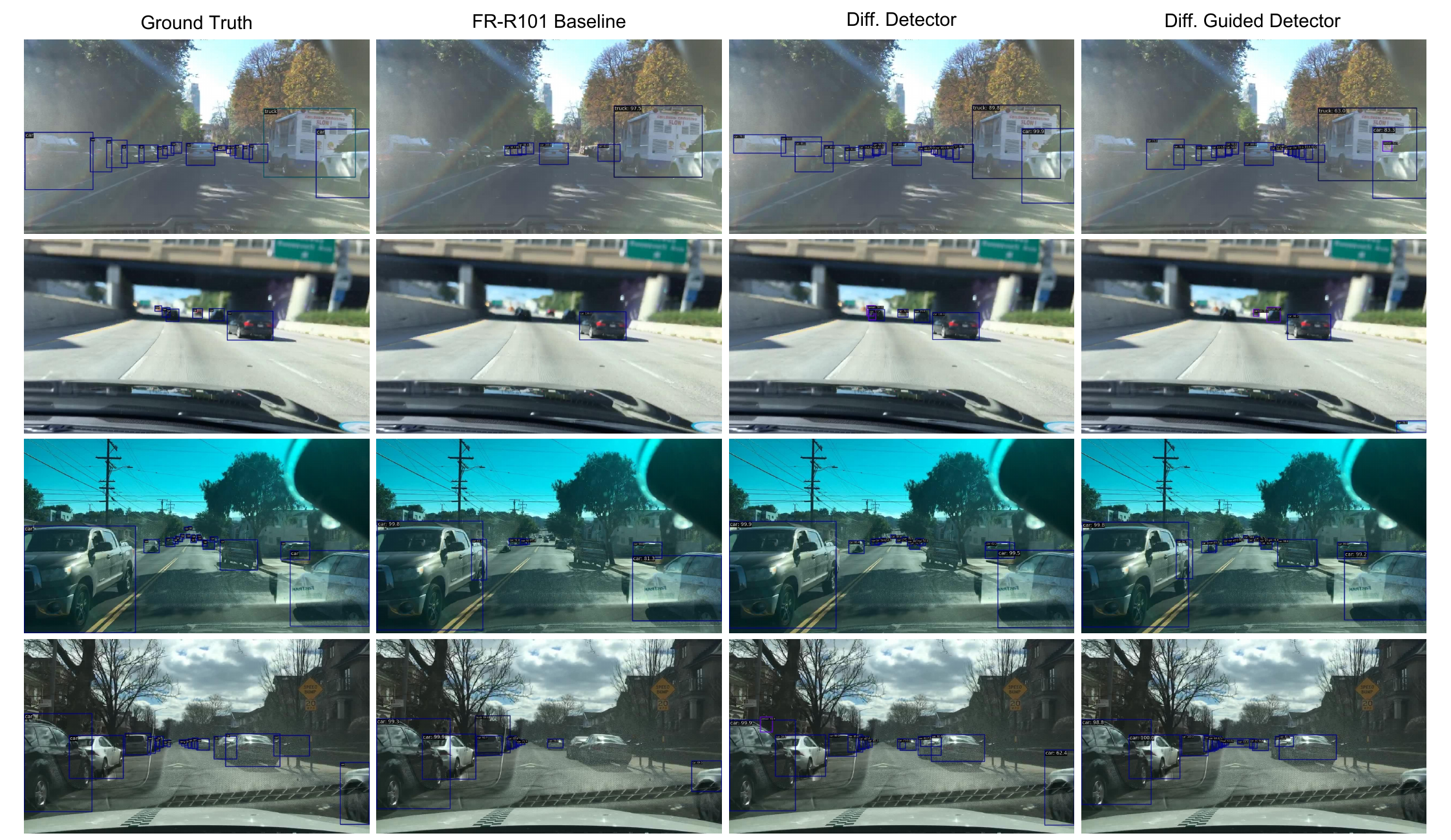}
    \caption{Qualitative prediction results on BDD100K.}
    \label{fig:vis_bdd100k}
\end{figure*}

\begin{figure*}[h]
    \centering
    \includegraphics[width=1.0\textwidth]{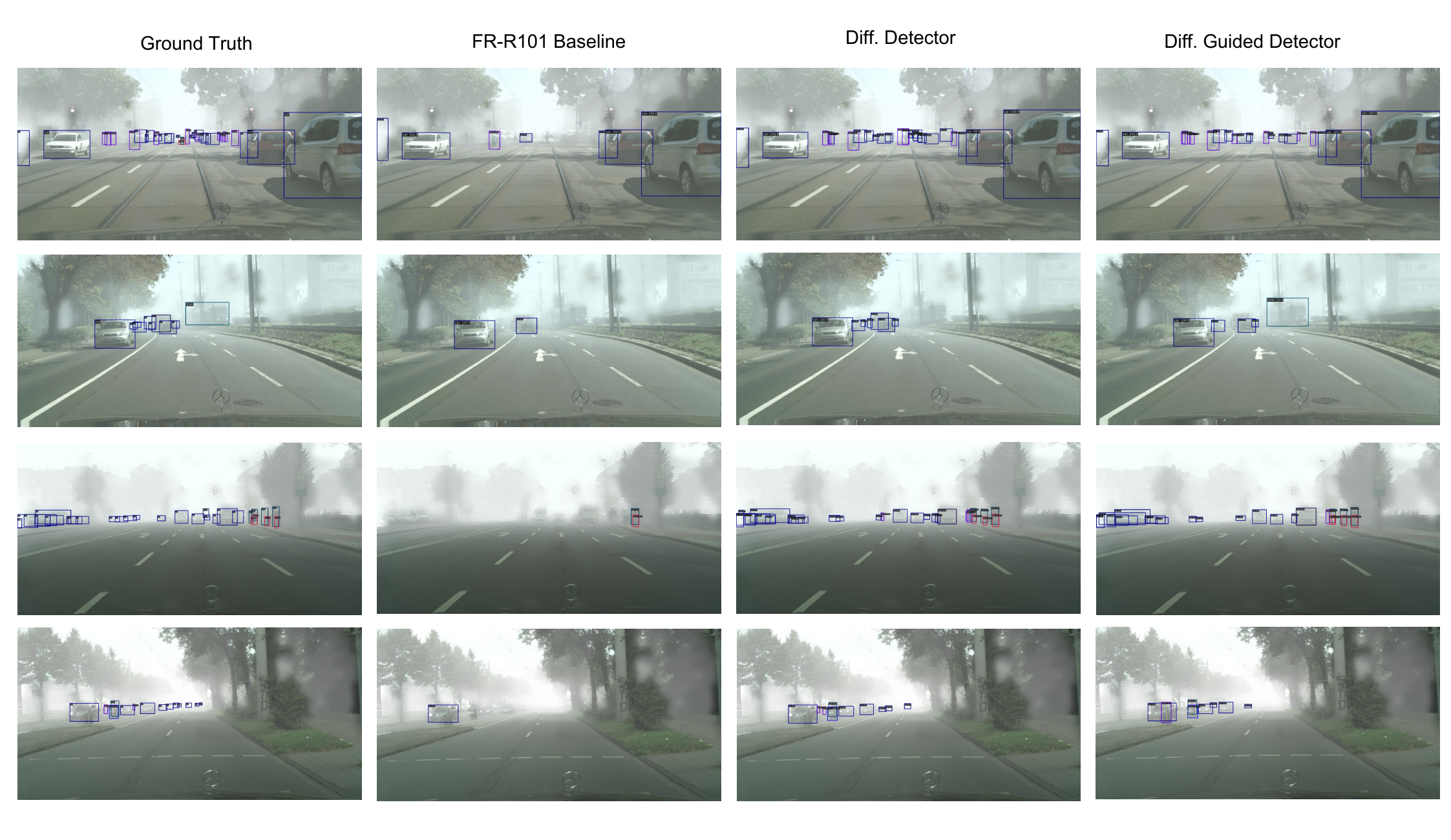}
    \caption{Qualitative prediction results on FoggyCityscapes.}
    \label{fig:vis_foggycityscapes}
\end{figure*}

\begin{figure*}[h]
    \centering
    \includegraphics[width=1.0\textwidth]{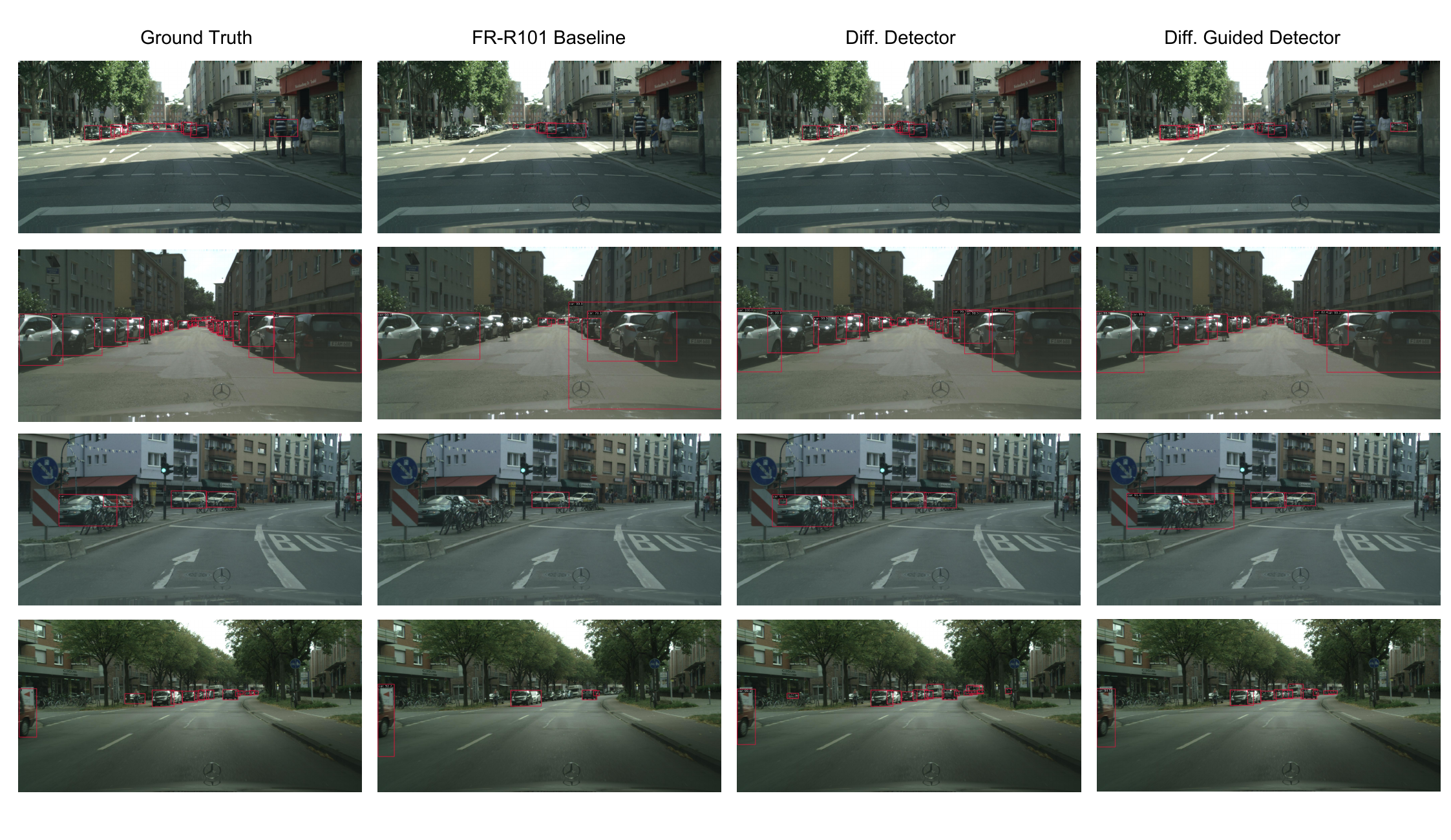}
    \caption{Qualitative prediction results on Cityscapes (Car).}
    \label{fig:vis_cityscapes(car)}
\end{figure*}

\begin{figure*}[h]
    \centering
    \includegraphics[width=1.0\textwidth]{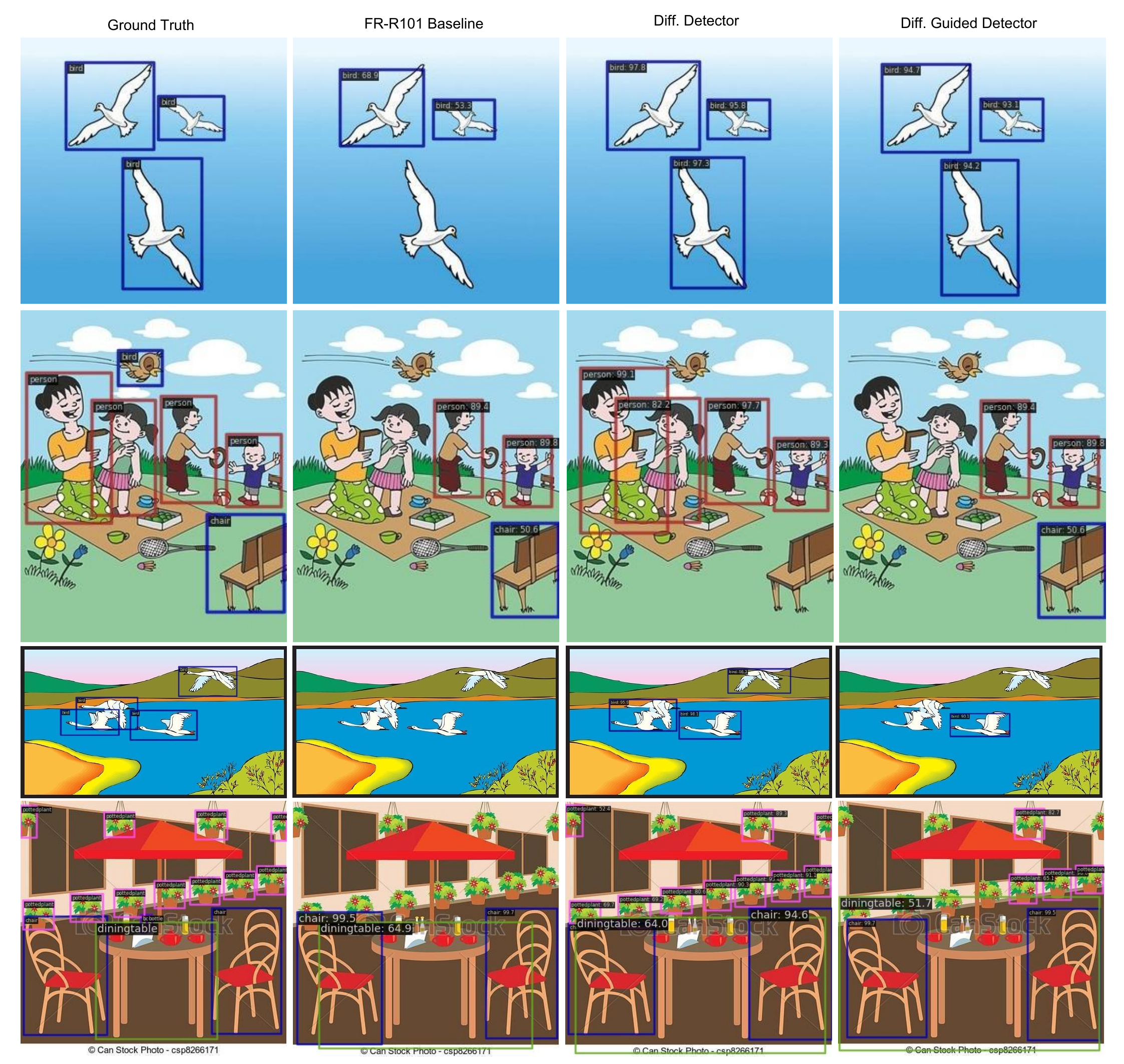}
    \caption{Qualitative prediction results on Clipart.}
    \label{fig:vis_clipart}
\end{figure*}

\begin{figure*}[h]
    \centering
    \includegraphics[width=1.0\textwidth]{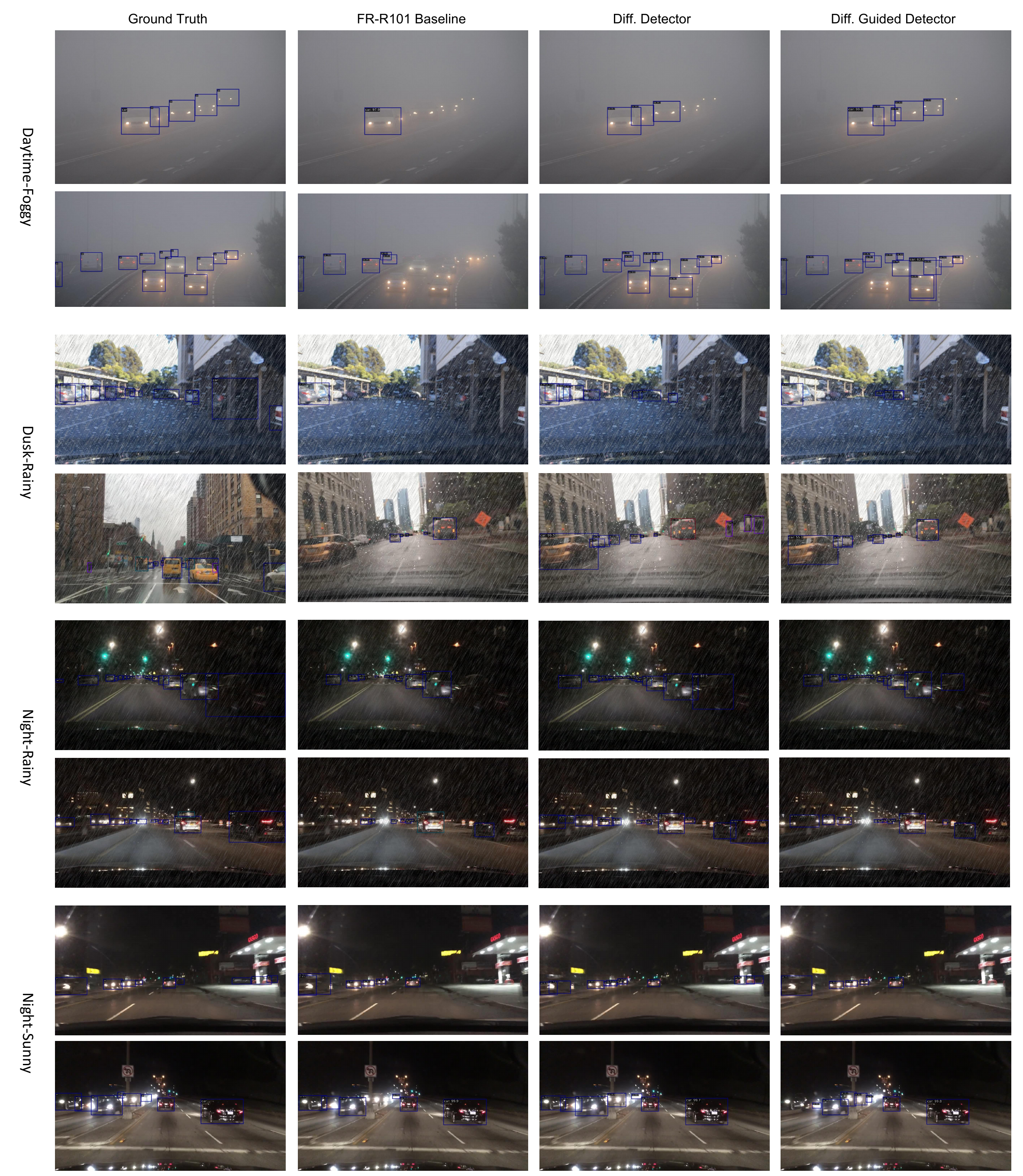}
    \caption{Qualitative prediction results on Diverse Weather Benchmark.}
    \label{fig:vis_dwd}
\end{figure*}

\begin{figure*}[h]
    \centering
    \includegraphics[width=1.0\textwidth]{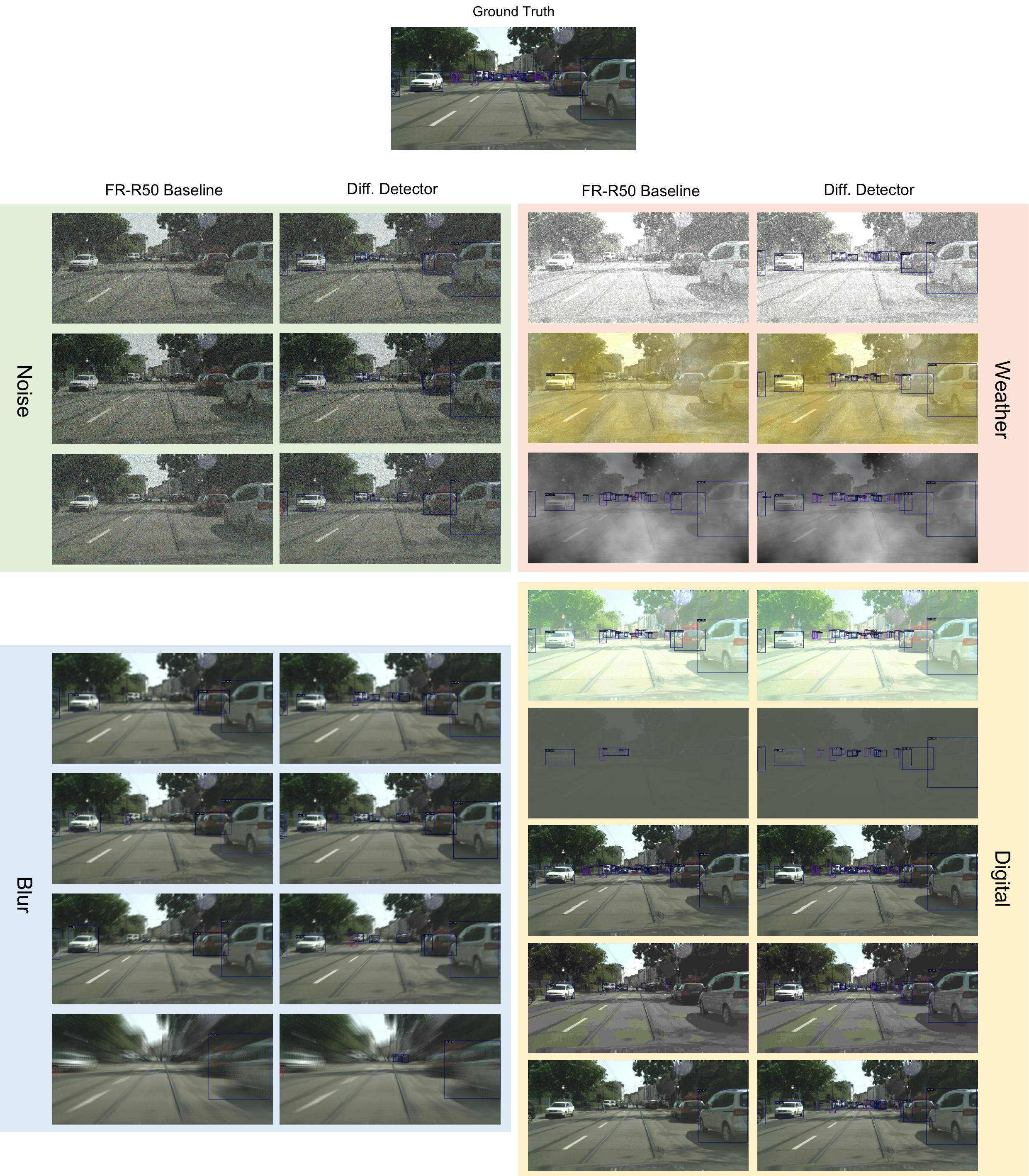}
    \caption{Qualitative prediction results on Corruption benchmark, showing detection results under 15 different corruption types (noise, blur, weather, and digital) at maximum severity level.}
    \label{fig:vis_cityscapes-c}
\end{figure*}

\noindent\textbf{Visualization Results:}
As shown in Fig.~\ref{fig:vis_bdd100k},~\ref{fig:vis_foggycityscapes},~\ref{fig:vis_cityscapes(car)},~\ref{fig:vis_clipart},~\ref{fig:vis_dwd}, and~\ref{fig:vis_cityscapes-c}, our proposed methods demonstrate superior detection performance across various challenging scenarios. Compared to the baseline detector, both Diff. Detector and Diff. Guided Detector achieve more comprehensive detection results, successfully identifying objects under different conditions such as varying scales, weather conditions, lighting variations, and artistic styles. These qualitative results consistently validate the effectiveness of our proposed diffusion-based framework in improving detection generalization across different domains.

\clearpage

\end{document}